\documentclass[11pt,a4paper,authoryear]{elsarticle} 
\usepackage[total={6.4in, 9.5in}]{geometry}

\usepackage{amssymb}

\usepackage{mathrsfs}   
\usepackage{amssymb}
\usepackage{optidef}
\usepackage{multirow}
\usepackage{hyperref}
\usepackage{placeins}
\usepackage[ruled,vlined,linesnumbered]{algorithm2e}
\usepackage[normalem]{ulem}
\useunder{\uline}{\ul}{}

\usepackage{xcolor}
\usepackage{cleveref}

\journal{Applied Soft Computing}

\begin{document}

\begin{frontmatter}


\title{Random-Key Algorithms for Optimizing Integrated Operating Room Scheduling}

\author[1,2]{Bruno Salezze Vieira\corref{cor1}}
\ead{bsvieira@unifesp.br}

\author[1,2]{Eduardo Machado Silva}
\ead{machado.silva@unesp.br}

\author[1]{Antônio Augusto Chaves}
\ead{antonio.chaves@unifesp.br}

\cortext[cor1]{Corresponding author}

\address[1]{Federal University of S\~ao Paulo (UNIFESP), S\~ao Jos\'e dos Campos, Brazil}
\address[2]{Aeronautics Institute of Technology (ITA), S{\~a}o Jos{\'e} dos Campos, Brazil}






\begin{abstract}

Efficient surgery room scheduling is essential for hospital efficiency, patient satisfaction, and resource utilization. This study addresses this challenge by introducing a novel concept of Random-Key Optimizer (RKO), rigorously tested on literature and new, real-world inspired instances. Our combinatorial optimization problem incorporates multi-room scheduling, equipment scheduling, and complex availability constraints for rooms, patients, and surgeons, facilitating rescheduling and enhancing operational flexibility. The RKO approach represents solutions as points in a continuous space, which are then mapped in the problem solution space via a deterministic function known as a decoder. The core idea is to operate metaheuristics and heuristics in the random-key space, unaware of the original solution space. We design the Biased Random-Key Genetic Algorithm with $Q$-Learning, Simulated Annealing, and Iterated Local Search for use within an RKO framework, employing a single decoder function. The proposed metaheuristics are complemented by lower-bound formulations, providing optimal gaps for evaluating the effectiveness of the heuristic results. Our results demonstrate significant lower and upper bounds improvements for the literature instances, notably proving one optimal result. Furthermore, the best-proposed metaheuristic efficiently generates schedules for the newly introduced instances, even in highly constrained scenarios. This research offers valuable insights and practical solutions for improving surgery scheduling processes, offering tangible benefits to hospitals by optimising resource allocation, reducing patient wait times, and enhancing overall operational efficiency.



\end{abstract}


\end{frontmatter}

\section{Introduction} \label{sec:intro}

Surgeries play a substantial role in hospital management. Their effective utilisation reduces surgical service delivery costs, shortens surgical patient wait times, and increases patient admissions \citep{roshanaei2017colaborative}. To optimally manage the most relevant decisions and constraints related to it, the integrated surgery scheduling problem arises. It is a complex optimization problem that involves determining the optimal schedule for a set of surgical procedures, considering various constraints and objectives.

Multiple operating rooms (ORs) are available in a hospital for surgeries. The goal of integrated surgery scheduling is to allocate the available resting beds, ORs, Intensive Care Units (ICUs), and Post-Surgery Care Units (PSCUs) to schedule surgical procedures in a way that maximizes room efficiency, reduces patient waiting times, and ensures the best utilization of resources.

\begin{figure}[!ht]
    \centering
    \scalebox{0.1}{\includegraphics{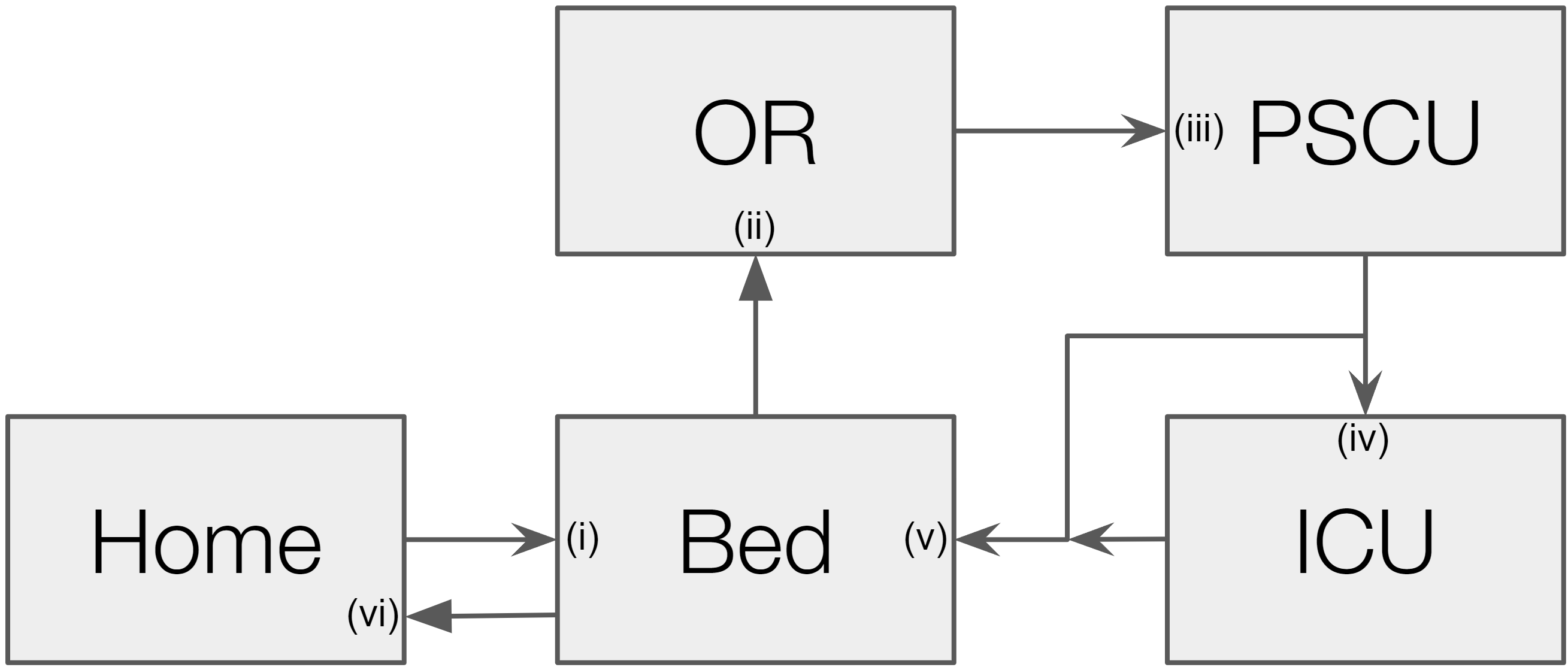}}
    \caption{Patient surgery flow.}
    \label{fig:patientFlow}
\end{figure}

From the patient's perspective, \autoref{fig:patientFlow} visually represents the surgery process. It begins with the patient arriving at the hospital and progressing through the following steps: ($i$) Resting room: Upon arrival, the patient is admitted and taken to a resting bed in the assessment area. Diagnostic tests and observation are conducted to assess the patient's medical condition; ($ii$) Operating Room (OR): Subsequently, the patient is taken to the OR to undergo the surgical procedure; ($iii$) Post-Surgery Care Unit (PSCU): After surgery, the patient is transferred to the PSCU for critical observation and anaesthesia recovery. This step ensures the patient's safe transition from the surgical procedure; ($iv$) Intensive Care Unit (ICU) (if applicable): Depending on the surgery's risk level or the patient's condition, there might be a scheduled transfer to the ICU for specialized care and monitoring; ($v$) Recovery Room: Following the critical observation and ICU (if required), the patient is taken back to the same room from step ($i$) for a recovery period. During this phase, the patient receives focused care and support to aid their healing process. Once the medical team determines the patient is stable and ready to leave the hospital, they are discharged ($vi$) with relevant post-surgery instructions and prescriptions, if necessary.

Considering this process, the problem we refer to as the Integrated Operating Room Scheduling Problem (IORSP) involves managing a set of surgical procedures, surgeons, operating rooms, and necessary equipment. A solution is to assign each patient to a suitable operating room based on predefined time slots at different stages within the previously outlined workflow. This allocation aims to schedule pending surgeries to minimize the total execution time, a metric referred to as makespan. This optimization problem takes into account a range of critical factors and constraints, which include:

\begin{itemize}
    \item Surgeon availability: The scheduling process must account for the availability and preferences of surgeons. Surgeons often have different specialities and expertise, and their schedules must be synchronized with the allocated ORs. In this case study, each surgery already has a surgeon selected for it;
    \item Operating room allocation: The problem entails allocating surgical procedures to the available operating rooms (ORs), which are only accessible during business hours, to minimize idle time and maximize utilization. Additionally, certain surgeries may necessitate specific equipment or specialized facilities unique to certain ORs, which adds complexity to the allocation process;
    \item Surgery duration: Each surgical procedure has an estimated duration, which must be considered when scheduling surgeries sequentially. Accurate estimation of surgery duration is crucial for avoiding delays and conflicts. The patient's time in all surgery steps must be accounted for when scheduling the multiple rooms;
    \item Equipment availability: The availability of necessary equipment for different kinds of surgeries, such as ophthalmic and brain microscopes or video/endoscopy racks, is considered during the scheduling process. Ensuring that all required resources are available at the right time is essential for smooth operations;
    \item Time constraints: The scheduling problem also consider time constraints, such as surgeon availability, business hours, moving time between rooms, and room maintenance post-use.
\end{itemize}

In our literature review, building upon the research of \cite{xiang2015antColoorsp}, \cite{burdett2018integFJSS}, and \cite{vali2022aplicFJSS}, we establish that the Integrated Operating Room Scheduling Problem (IORSP) can be effectively conceptualized and addressed as a variant of the Flexible Job Shop Problem (FJSP) \citep{brucker1990firstfjss}. This research has been guided by the collaboration with a non-profit hospital that has unities in several cities in Brazil, basing our approach on a real-world case. This paper reviews existing literature alongside a comparative analysis with analogous scenarios and methodologies.

In this paper, we propose three metaheuristics using the Random-Key Optimizer (RKO) concept \citep{chaves2024randomkeygraspcombinatorialoptimization} to solve the IORSP: Biased Random Key Genetic Algorithm with $Q$-Learning (BRKGA-QL) \citep{chaves2021brkgaql}, Simulated Annealing (SA) \citep{Kirkpatrick1983}, and Iterated Local Search (ILS) \citep{lourenco2003iterated}. The BRKGA-QL has demonstrated efficacy without necessitating parameter fine-tuning, which is crucial for real-case hospital applications. Furthermore, we introduce two mathematical models for relaxed cases, enabling more efficient computation of lower bounds compared to a complete formulation \citep{chaudhry2016fjssreview}. Additionally, we tested available literature instances, adapting our methods to solve them successfully and comparing our results with existing literature methods. Finally, we designed, tested, and made a set of 20 instances based on our case study available online.

We can highlight our scientific contributions as:

\begin{itemize}
    \item Proposal of two lower bound formulations that have small implementation complexity and yield better lower bounds than previous literature approaches;
    \item Investigation of flexible time constraints for each restricted schedule, where each room/surgeon's working hours can be delimited as available or not by the minute (one minute was our time discretization approach, but the user can change it);
    \item Introduction of a novel concept of a random-key optimizer with three metaheuristics using the same decoder process;
    \item Development of a parameter-less metaheuristic that performs better than tuned ones and literature algorithms for similar hospital cases.
    \item Validation of our methods against a similar literature case and demonstration of better results than previous ones;
    \item Complete modelling of a real-case scenario by procedurally generating 20 instances and making them available in a public repository;
    \item Demonstration of possible rescheduling using the same input data approach;
\end{itemize}

The remainder of the paper is structured as follows. In Section \ref{sec:litReview}, we provide a literature review of the most similar works. In Section \ref{sec:prob_math}, we define the problem description. Section \ref{rko} details our proposed metaheuristics using the random-key concept, decoder, and reinforcement learning component. In Section \ref{sec:compExp}, we present our experimental data, results, and analyses. In Section \ref{sec:conclusions}, we further discuss the results and conclude the paper. Our mathematical formulations to compute lower bounds are shown in \ref{sec:relaxed_model_bed} and \ref{sec:relaxed_model_or}.


\section{Literature review} \label{sec:litReview}

The modeling of surgery scheduling problems commenced relatively early in deterministic models literature, among the first works we have \cite{ozkarahan1995alloc}. The authors developed software with multiple allocation models to schedule nurses, surgeon blocks, and surgeries and, followed by \cite{dexter1999binpack}, studied bin packing algorithms and fuzzy constraints in operating room management and other works that followed.

Reviewing the literature, given how diverse hospital management is worldwide, the cases modelled after real-world cases rarely match each other, but they have enough similarities as we list them. Most works model the problem only scheduling the operating rooms \citep{abdelrasol2014orspSurv}. Among the works most similar to our approach, we have first the works of \cite{fei2009tatical}, \cite{fei2010planning}, and \cite{liu2011heurORSP} that modelled OR allocation with five days of planning considering working hours and surgeon scheduling and solved using column generation and dynamic programming heuristics for procedurally generated instances and a Belgian university hospital case study. \cite{molina2015integrated} proposed a case that may allocate an assistant surgeon to shorter surgery times. They proposed and implemented a constructive heuristic to solve it.

\cite{aringhieri2015twolevel} modelled a case from the San Martino University Hospital, addressing simultaneous OR and recovery bed scheduling for a one-week planning horizon. Their model considers weekend stays without operations and various surgery specialities and is solved using a two-stage metaheuristic (assign and then schedule). \cite{landa2016hybrid} extended the same case study into a stochastic variant, aiming to maximize OR utilisation and minimize overtime costs, and solved it using a hybrid simulation-based algorithm.

\cite{xiang2015antColoorsp} treated the problem as an FJSP variant, scheduling pre and post-surgery rooms, nurses, and anaesthetists individually, similarly to our approach. They solved it with an Ant Colony metaheuristic. \cite{marques2015orElective} modelled a case of Lisbon's public hospital, allocating only ORs in a one-week planning horizon. The bi-objective problem maximizes the number of surgeries and room occupations with surgeon specialities and is solved with a bi-objective evolutionary metaheuristic.

\cite{duran2017prioriLists} modelled a case of a Chilean public hospital and developed two models for scheduling interventions on top of an existing OR schedule over a defined period that satisfies patient priority criteria. \cite{dellaert2017vnsSTP} modelled a case of a Dutch cardiothoracic centre for a 4-week planning horizon, considering the allocation of beds, ORs, and ICUs and maximizing the number of scheduled surgeries. The authors proposed a mathematical model and a Variable Neighbourhood Search (VNS) metaheuristic that found better results than the model solved with CPLEX in all cases. \cite{siqueira2018longInt} solved a stochastic model for a long-term plan from a case study of the Brazilian National Institute of Traumatology and Orthopedics, which considers OR and recovery ward allocations with downstream constraints, using simulations and optimal action-taking.

\cite{burdett2018integFJSS} modelled a case of a university and college-affiliated teaching hospital in Brisbane, Australia, as an FJSP considering bed, OR, PSCU, and ICU allocation. They presented and made available $24$ generated instances, and to solve it, they proposed some initial solution heuristics and a Hybrid Simulated Annealing (HSA) as its main algorithm. \cite{hamid2019teamMemb} modelled a public hospital case in Tehran as a multi-objective problem. They schedule ORs considering patient priorities and surgical team members' decision-making styles to improve the surgical teams' compatibility level, minimizing the total cost, overtime OR utilisation, and maximizing team consistency. They presented a mathematical model and implemented two metaheuristics to obtain better solutions: a Non-dominated Sorting Genetic Algorithm (NSGA-II) and a Multi-Objective Particle Swarm Optimisation (MOPSO). 

\cite{roshanaei2020branchCheck} proposed Branch-and-check methods for operating room planning and scheduling with surgeon specialities for patient surgeon assignment and priorities. Their exact methods outperformed conventional integer formulation approaches by a large margin using the benchmark instances of \cite{marques2015orElective} and a procedurally generated one. \cite{zhu2020dyn3Stage} solved a case study of an affiliated hospital of the University of Science and Technology of China, maximizing OR utilisation on a one-week planning horizon, OR to surgery to surgeon assignment, and patient priorities. The authors presented a mathematical formulation and two hybrid metaheuristics to obtain better solutions. \cite{schneider2020downRes} studied the surgery groups scheduling considering ORs and downstream beds operating with different available hours with a 15-day planning horizon, modelled of the Leiden University Medical Center, proposed a single step integer model that maximizes OR usage and minimises bed usage variation with weights, they also tested the \cite{vanEssen2014groupSched} instances lowering the amounts of required beds upon the previous results.

More recently, \cite{lin2021orspArtBee} solved a case of surgery to OR assignment, five days planning horizon and minimizing the operation costs and overtime usage, provided an improved mathematical model of \cite{fei2009tatical} and a metaheuristic Artificial Bee Colony (ABC) that obtained better solutions for larger instances. \cite{park2021orspPref} modelled a case of a Korean university hospital considering surgeons’ preferences and cooperative operations, in which surgery can have multiple surgeons cooperatively sequentially to reduce its execution time, and the co-oped surgeries have a lower total execution time. This feature helps to minimize the total overtime and the number of ORs used. A mathematical model and a metaheuristic were proposed to solve it.

\autoref{tab:biblReview} comprises the similarities found during our literature review for the modelled real-world case studies, the case study they were based on, if it was solved with an exact (E) or heuristic (H) method, objective function focus, and the problem attributes: surgery to surgeon assignment (SSA), multi-room scheduling (MRS), time slots allocation (TS), patient priorities (PP), surgeon specialities (SE), a limited planning horizon (PH), overtime costs (OT), and other resources scheduling (ORS) like nurses, anaesthetists or equipment.

\begin{table}[ht]
\centering
\caption{Case study comparisons with similar literature works.}
\label{tab:biblReview}
\scalebox{0.575}{
\begin{tabular}{l|l|ccl|cccccccl}
Paper                                                   & Case study                           & E & H & Objective Function & SSA & MRS & TS & PP & SE & PH & OT & ORS \\ \hline
\cite{fei2009tatical, fei2010planning, liu2011heurORSP} & Belgian university hospital          & X & X & Costs              &     &     & X  &    &    & X  & X  &    \\
\cite{aringhieri2015twolevel, landa2016hybrid}          & San Martino university hospital      & X & X & Costs              &     &     & X  &    & X  & X  &    &    \\
\cite{marques2015orElective}                            & Public hospital in Lisbon            &   & X & Usage              &     &     & X  & X  & X  & X  &    &    \\
\cite{xiang2015antColoorsp}                             & -                                    &   & X & Makespan           & X   & X   & X  &    & X  &    & X  & X  \\
\cite{duran2017prioriLists}                             & Public hospital in Chile             &   & X & Usage              & X   &     &    & X  & X  & X  & X  &    \\
\cite{dellaert2017vnsSTP}                               & Dutch cardiothoracic center          & X & X & Usage              &     & X   &    &    &    & X  &    &    \\
\cite{burdett2018integFJSS}                             & Hospital in Brisbane, Qld, Australia & X & X & Makespan           &     & X   & X  &    &    &    &    &    \\
\cite{hamid2019teamMemb}                                & Public hospital in Tehran            & X & X & Usage and MSWT     &     &     & X  & X  &    & X  & X  &    \\
\cite{roshanaei2020branchCheck}                         & Public hospital in Lisbon            & X &   & Usage              & X   &     & X  &    & X  & X  &    &    \\
\cite{zhu2020dyn3Stage}                                 & Affiliated one of UST of China       &   & X & Costs              & X   &     &    &    & X  & X  & X  &    \\
\cite{schneider2020downRes}                             & Leiden University Medical Center     & X & X & Usage              &     &     & X  &    & X  & X  & X  &    \\
\cite{yazdi2020elective}                                & Medium-sized Norwegian Hospital      & X &   & \#Surgeries        & X   & X   & X  &    &    & X  &    &    \\
\cite{akbarzadeh2020divHeur}                            & Sina Hospital (Tehran, Iran)         & X & X & Costs              &     &     & X  &    &    & X  &    & X  \\
\cite{park2021orspPref}                                 & Korean university hospital           & X & X & Active ORs and OT  &     &     & X  &    &    &    & X  & X  \\
This Work                                               & Non profit hospital in Brazil        & X & X & Makespan           &     & X   & X  & X  &    &    &    & X 
\end{tabular}}
\end{table}

For more comprehensive reviews of ORSP models, including models with uncertainty, stochastic, and mixed approaches, we advise the works of \cite{abdelrasol2014orspSurv}. \cite{vanriet2015tradElectiveReview} presented a review focused on the trade-offs in operating room planning for electives and emergencies, and \cite{rahimi2021review} accomplished analysis and classified solution approaches similarly to \cite{abdelrasol2014orspSurv}, with deterministic and probabilistic approaches, performance metrics and its trends throughout time.

\section{Problem definition} \label{sec:prob_math}

Given the literature review and problem characteristics, we model the IORSP as an FJSP variant. The FJSP is usually described as having a set of jobs $K$ to be processed by the machines of set $R$, each job consists of a sequenced set $T_k$ of tasks, and each task $t$ has to be performed sequentially to complete the job. For each job $k$, $k\in K$, the execution of task $t$, $t\in T_{k}$, requires one machine $r$ out of a set of given machines $R_{t}$, $R_{t}\subset R$, that execute task $t$. For task $t$ running on machine $r$, $r \in R_{t}$, the task time is $\gamma_t^{d}$, the setup time is $\gamma_t^{m}$ and cool-down time is $\gamma_t^{c}$. When a task $t$ is complete, but the allocated machine $r \in R_{Next(t)}$ is not immediately available, the start of task $Next(t)$ is delayed. This delay is called blocking, with a blocking limit per task $\phi_{t}$.

Moreover, the FJSP typically follows some assumptions as we do, as the machines and jobs are always available. Once started, an operation cannot be interrupted. There is no precedence among the tasks of different jobs; each task can be processed by only one machine at a time. As the objective function, FJSP usually minimizes the makespan, that is, the amount of time required to complete all jobs or to maximize the total of completed jobs given a planning horizon. 

We can translate the FJSP nomenclature to our problem, considering jobs as surgeries and machines as subjects (rooms, types of equipment, and surgeons). Each surgery consists of a sequence of tasks. Our problem has some particularities, like a non-empty initial schedule or simultaneous machines for the same task, as operations (tasks) require a surgeon, which may require some equipment and OR, i.e., different types of machines (subjects) allocated simultaneously. All $\gamma_t^{d}$, $\gamma_t^{m}$ and $\gamma_t^{c}$ are fixed for any particular subject, so they are called duration ($\gamma_t^{d}$), moving time ($\gamma_t^{m}$) and cleaning time ($\gamma_t^{c}$). The name blocking is a homonym in our work, and we have a fixed blocking limit $\Phi$ for all tasks of $15$ minutes. Each subject (machine) has its initial availability schedule (structure we call availability slots). As an FJSP variant, the surgery is a job that requires heterogeneous machines working on the same task during the same period to complete it. In our approach, we only tackled minimizing the makespan. We note that maximizing the number of surgeries tends to prioritize the shorter surgeries.

\autoref{tab:inputNotations} shows our modelled structures for our mathematical notations. We can divide them between static and dynamic structures. Among our static structures, the input data consists of a set of equipment $E_{f}$ for each equipment type $f$, sets $R^{bed}$ and $R^{or}$ representing the beds and operation rooms, respectively, a set of surgeries $K$, and each surgery $k$ to be scheduled has a set of tasks $T_k$. Each task $k$ has a set of required people $L_t$ (in our case study, this set primarily includes patients for all tasks and the assigned surgeon for each surgery), set of required equipment types $F_t$, set of compatible rooms $R_t$, expected duration time $\gamma_t^{d}$, moving time $\gamma_t^{m}$ and cleaning time $\gamma_t^{c}$. There is an initial set of availability slots for each one of our resources (person, equipment, or room).

\begin{table}[h]
\centering
\caption{Problem data notations}
\label{tab:inputNotations}
\scalebox{0.85}{
\begin{tabular}{l|l} \hline 
\textbf{Sets}     &           \\ 
$E_{f}$       & Equipment of type $f$       \\
$R^{or}$     & All OR rooms   \\
$R_{bed}$    & All bed rooms \\ 
$K$           & Surgeries               \\ 
$T_k$         & Tasks of surgery $k$        \\
$L_{t}$       & People allocated for task $t$    \\ 
$F_t$         & Equipment types required for task $t$   \\
$R_t$         & Compatible rooms for task $t$ \\ \hline
\textbf{Attributes}     &            \\ 
$\gamma_t^{d}$       & Duration time of task $t$  \\ 
$\gamma_t^{m}$       & Moving time of task $t$  \\ 
$\gamma_t^{c}$       & Cleaning time of task $t$  \\  \hline
\end{tabular}}
\end{table}

\begin{figure}[!ht]
    \centering
    \scalebox{0.20}{\includegraphics{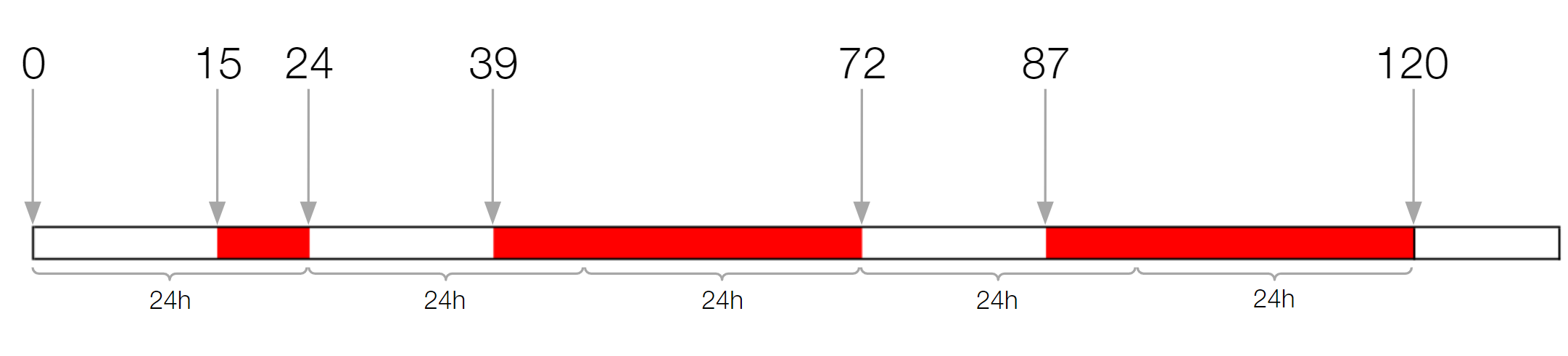}}
    \caption{Example of a set of availability slots (AS) for a subject.}
    \label{fig:bw_schedule}
\end{figure}

\autoref{fig:bw_schedule} illustrates the concept of availability slots (AS), a timeline for any subject with available and unavailable intervals. The numbers and arrows on the illustration point to the corresponding hour an interval starts or ends. In this example $W=\{ 0,15,24,39,72,87,120\}$, the subject is available for the first 15 hours, becomes unavailable for 9 hours (15 to 24), is available again for 15 hours (24 to 39), becomes unavailable for the next 33 hours (39 to 72), and so on. This AS example could be the availability of a surgeon, operating room, or other subjects. Mathematically, we define $W$ as an ordered set of time instants arranged in increasing order. The initial element signifies the first available time instant, while the subsequent elements represent an alternation of unavailable and available time instants. This data structure is used as an initial timeline availability and is dynamically adjusted per solution.

Furthermore, our modelling allows tackling the re-scheduling of surgeries. Surgery re-scheduling is crucial for effectively managing the IORSP due to the hospital environment's dynamic and unpredictable nature. It allows for flexibility and adaptability in emergencies, resource availability fluctuations, and changes in patient conditions. Re-scheduling ensures that urgent surgeries can be accommodated without significant disruptions, minimizes delays and wait times, and optimizes the use of resources. Additionally, it helps cope with cancellations and no-shows by filling gaps efficiently, maintaining high productivity levels, and ensuring timely surgical care. To implement surgery re-scheduling efficiently, we make use of the sets of availability slots with the previously fixed schedules for rooms, patients, equipment and surgeons as shown in detail in Section \ref{sec:surg_rescheduling}.

\ref{sec:relaxed_model_bed} and \ref{sec:relaxed_model_or} preset the relaxed models we propose and implement to compute lower bounds to minimize each instance's makespan. The resulting lower bounds are used to analyze the performance of the heuristics, and smaller gaps imply a strong performance of both proposals. Both models solve a relaxed case of the IORSP and exploit the fact that, for the case study and the literature case, the bed is reserved during the entire patient's stay in the hospital. It considers a best-case scenario execution for all surgeries and only allocates surgeries to rooms but does not sequence any task. All patients are treated without delays (blocking), disregarding room or surgeon-restricted schedules, sequencing of tasks, and equipment availability. Removing these constraints, the relaxed problem solved by both models is an allocation problem.

\section{Random-Key Optimizer} \label{rko}

The random-key representation was first proposed by \cite{bean1994rkga} for an extension of the genetic algorithm called the Random-key Genetic Algorithm (RKGA). This representation encodes a solution with random numbers in the interval $[0,1)$. The main idea is that the RKGA searches the random-key space as a surrogate for the original solution space. Points in the random-key space are mapped to points in the original solution space by a deterministic algorithm called the decoder. An advantage of this encoding is its robustness to problem structure, as it separates the solver (solution procedure) from the problem being solved. The connection between the solver and the problem is established through the decoder.

\begin{figure}[htbp]
    \centering
    \includegraphics[width=0.75\linewidth]{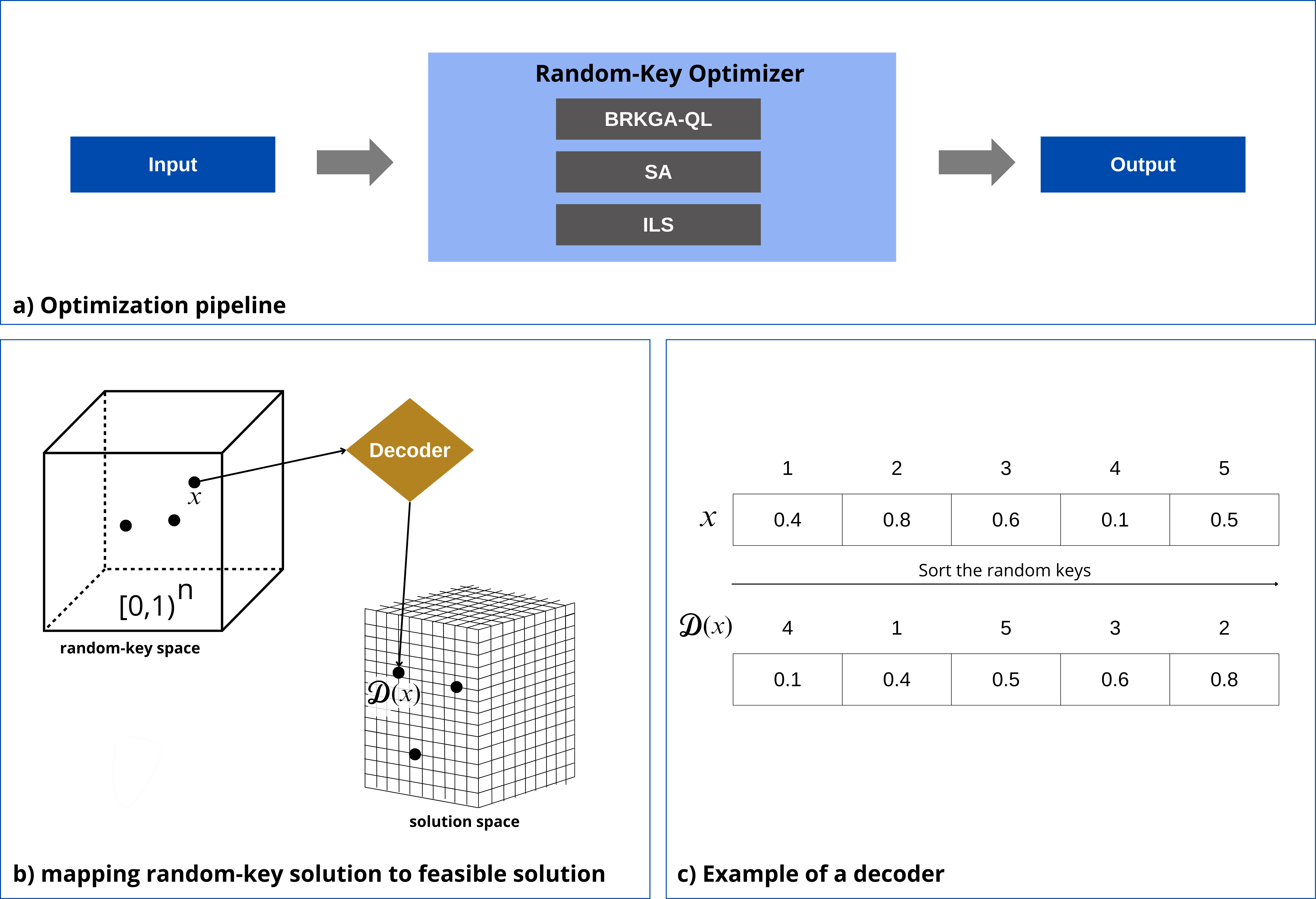}
    \caption{Random-Key Optimizer concept. Based on \cite{schuetz2022rko}.}
    \label{fig:rko}
\end{figure}

A Random-key Optimizer (RKO) is an optimization heuristic that solves specific problems by exploring the continuous random-key space $[0,1)^{n}$, where $n$ is the size of the random-key vector that encodes the problem solution. The pioneer of this approach is the work of \cite{bean1994rkga}, which utilizes a genetic algorithm to perform searches in the solution space for various optimization problems. In \cite{goncalves2011brkga}, a variation of the RKGA called Biased RKGA (BRKGA) is proposed and evaluated for a wide range of optimization problems. Later, the RKO framework was employed by \cite{schuetz2022rko} for robot motion planning, using the BRKGA and an extension of Simulated Annealing to search the random-key space. In \cite{mangussi2023meta}, the RKO is implemented using Simulated Annealing, Iterated Local Search, and Variable Neighbourhood Search for the tree hub location problem. Recently, \cite{chaves2024randomkeygraspcombinatorialoptimization} proposed an RKO considering the GRASP metaheuristic and evaluates it for the travelling salesman problem, tree hub location problem, Steiner triple covering problem, node capacitated graph partitioning problem, and job sequencing and tool switching problem. \autoref{fig:rko} presents the RKO concept where \(\mathscr{D}\) represents the specific decoder algorithm. 

The optimization process is shown in \autoref{fig:rko}(a). This process takes as input an instance of a combinatorial optimization problem and returns the best solution found, guided by a specified metaheuristic. \autoref{fig:rko}(b) illustrates the mapping schema that connects the random-key representation to the solution space through a problem-specific decoder. After this transformation, the quality of the solutions can be evaluated. Finally, \autoref{fig:rko}(c) presents an example of a decoder process for a scheduling-based problem. In this case, each vector position represents a task or characteristic of the optimization problem. The decoder works by sorting the random keys, where the sorted indices correspond to a potential solution obtained by arranging the tasks in the specific order dictated by the sorted keys.

In the remainder of this section, we introduce the RKO framework. We begin by discussing its key components (subsection \ref{rkocomponents}), followed by an overview of the metaheuristics utilized in the framework. Specifically, we present the BRKGA-QL (subsection \ref{sec:BRKGA-QL}), Simulated Annealing (subsection \ref{secSA}), and Iterated Local Search algorithms (subsection \ref{secILS}), which operate on the random-key vectors within the proposed RKO framework.

\subsection{RKO Components}\label{rkocomponents}

The RKO components are procedures embedded within the framework's random-key space that support the search process's metaheuristics balance between diversification and intensification. These components include shaking, blending, and local search performed by the Randomized Variable Neighbourhood Descent (RVND) method on the random-key vectors. Each of these components is described in detail below.

\subsubsection{Shaking}

 The shaking method was inspired by the approach proposed by \cite{ANDRADE201967}. The method modifies random-key values by applying random modifications considering four distinct neighbourhood moves. A perturbation rate \( \beta \) is employed. This value is randomly generated within a specified interval \([\beta_{min}, \beta_{max}]\), which should be defined according to the specific metaheuristic approach being used. The four movements are:
\begin{itemize}
    \item \textit{Swap}: Swap the positions of two randomly selected random keys \( i \) and \( j \).
    \item \textit{Swap Neighbor}: Swap the position of a randomly selected random key \( i \) with its neighboring key \( i + 1 \).
    \item \textit{Mirror}: Change the value of a randomly selected random key \( i \) with its complementary value.
    \item \textit{Random}: Assigns a new random value within the interval \([0, 1)\) to a randomly selected random key \( i \).
\end{itemize}

Algorithm \ref{alg:shaking} described the shaking procedure. First, a shaking rate \( \beta \) is randomly generated within the interval \([\beta_{min}, \beta_{max}]\), determining the number of perturbations to be applied, specifically \( \beta \times n \), where \( n \) is the length of the random-key vector \( A \). For each perturbation, a random shaking move is selected from four options: a random move, a mirror move, a swap move, or a swap neighbor move. The selected move is then applied to the vector \( A \). After all perturbations are performed, the modified vector is returned as the output. This vector is then decoded during the metaheuristics search process.

\begin{algorithm}
\caption{Shaking method}
\label{alg:shaking}
\SetAlgoLined
\KwIn{Random-key vector $A$, $\beta_{min}$, $\beta_{max}$}
\KwOut{Changed random-key vector $A$}
Generate shaking rate $\beta$ randomly within the interval $[\beta_{min}, \beta_{max}]$\;
\For{$k \gets 1$ \textbf{to} $\beta \times n$}{
    Randomly select one shaking move $m$ from $\{1,2,3,4\}$\;
    \Switch{$m$}{
        \Case{$1$}{Apply Random move in $A$\;}
        \Case{$2$}{Apply Invert move in $A$\;}
        \Case{$3$}{Apply Swap move in $A$\;}
        \Case{$4$}{Apply Swap Neighbor move in $A$\;}
    }
}
\Return $A$\;
\end{algorithm}

\subsubsection{Blending}
\label{sec:blending}

The blending method extends the uniform crossover (UX) concept proposed by \cite{davis1991handbook} by introducing stochastic elements to create a new random-key vector. The algorithm combines two solutions, \( A^a \) and \( A^b \), to generate a new solution \( A^c \). For each position \( i \) in the vector, a random decision is made based on a probability \( \rho \) to inherit the corresponding key from either \( A^a \) or \( A^b \). The algorithm introduces a parameter $\mathit{factor}$, which modulates the contribution of \( A^b \). Specifically, when $\mathit{factor}$ = 1, the original key from \( A^b \) is used, and when $\mathit{factor}$ = -1, the complement \( (1.0 - A^b_i) \) is considered. Additionally, with a small probability \( \mu \), the algorithm generates a new random value within the interval \([0,1)\), further diversifying the resulting vector \( A^c \). The algorithm's pseudocode is presented in Algorithm \ref{alg:blending}.

\begin{algorithm}
\caption{Blending method}
\label{alg:blending}
\SetAlgoLined
\KwIn{Random-key vector $A^a$, Random-key vector $A^b$, $\mathit{factor}$, $\rho$, $\mu$}
\KwOut{New random-key vector $A^c$}
\For{$i \gets 1$ \textbf{to} $n$}{
    \If{$\textrm{{\fontfamily{pcr}\selectfont UnifRand}}(0,1) < \mu$}{
        $A^c_i \leftarrow \textrm{{\fontfamily{pcr}\selectfont UnifRand}}(0,1)$\;
    }
    \Else{
        \If{$\textrm{{\fontfamily{pcr}\selectfont UnifRand}}(0,1) < \rho$}{
            $A^c_i \leftarrow A^a_i$
        }
        \Else{
            \If{$\mathit{factor}=1$}{$A^c_i \leftarrow A^b_i$}
            \If{$\mathit{factor}=-1$}{$A^c_i \leftarrow 1.0 - A^b_i$}
        }        
    }
}
\Return $A^c$\;
\end{algorithm}

\subsubsection{Randomized Variable Neighbourhood Descent}
\label{sec:rvnd}

The Variable Neighbourhood Descent (VND) was proposed by \cite{MLADENOVIC19971097} and extended later for various optimization problems. The VND consists of a finite set of pre-selected neighbourhood structures denoted by $N_k$ for $k = 1, \ldots, k_{max}$, where $N_k(A)$ represents the set of solutions in the $k$-th neighborhood of a random-key vector $A$. While standard local search heuristics typically employ a single neighborhood structure, VND utilizes multiple structures to enhance the search process. Key considerations for applying VND include determining which neighborhood structures to use and their sequence and selecting an appropriate search strategy for switching between neighborhoods. Later, \cite{subramanian2010parallel} proposed the RVND. RVND randomly selects the neighborhood heuristic order to be applied in each iteration. RVND efficiently explores diverse solution spaces and can be applied to random-key spaces. Users can implement classic heuristics for the specific problem and encode the locally optimal solution into the random-key vector after the search process. Alternatively, users can implement random-key neighborhoods independent of the specific problem, using the decoder to converge towards better solutions iteratively. 

Algorithm \ref{alg:RVND} displays the RVND pseudo-code. Given an initial solution \( A \), the algorithm begins by initializing a Neighborhood List ($NL$). While the $NL$ is not empty, a neighborhood \( \mathcal{N}^i \) is selected randomly from it, and the best neighbor \( A' \) within \( \mathcal{N}^i \) is identified. If the objective function value \( \delta_{\mathcal{D}(A')} \) improves upon the current solution \( \delta_{\mathcal{D}(A)} \), the current solution \( A \) is updated to \( A' \), and the $NL$ is reset. If no improvement is found, the selected neighborhood \( \mathcal{N}^i \) is removed from the $NL$. The process repeats until all neighborhoods have been explored without finding a better solution. The algorithm then returns the best solution found.


\begin{algorithm}[htbp]
    \KwIn{$A$}
    \KwOut{The best solution in the neighbourhoods.}
    Initialise the Neighbourhood List ($NL$)\;
    \While{$NL \neq 0$}{
        Choose a neighbourhood $\mathcal{N}^i \in NL$ at random\;
        Find the best neighbour $A'$ of $A \in \mathcal{N}^i$\;
        \If{$\delta_{\mathcal{D}(A')} < \delta_{\mathcal{D}(A)}$}{
            $A \leftarrow A'$\;
            Restart $NL$\;
        }
        \Else{
            Remove $\mathcal{N}^i$ from the $NL$ \;
        }
    }
    \Return $A$
\caption{RVND} \label{alg:RVND}
\end{algorithm}

Next, we introduce four problem-independent local search heuristics designed to operate within the random-key space. These heuristics employ distinct neighborhood structures for the RVND algorithm, specifically used to identify the best neighbors as described in line 4 of Algorithm \ref{alg:RVND}. The neighborhood structures include Swap LS, Mirror LS, Farey LS, and Nelder-Mead LS.

\subsubsection{Swap Local Search}

The Swap local Search focuses on interchanging two values within the random-key
vector. The local search procedure considering this structure is outlined in Algorithm \ref{alg:swap}. The algorithm begins by defining a vector \( RK \) with random order for the random-key indices and initialises the best solution found, \( A^{best} \), to the current solution \( A \). It then iterates over all pairs of indices \( i \) and \( j \) (with \( j > i \)) in the random-key vector. For each pair, it swaps the value of the random keys at indices \( RK_i \) and \( RK_j \) in \( A \). If the resulting solution has a better objective function value than \( A^{best} \), it updates \( A^{best} \) to the new solution. If not, it reverts \( A \) to the previous best solution. The process continues until all pairs have been considered. The algorithm returns \( A^{best} \) as the best random-key vector found in the neighborhood.

\begin{algorithm}[h]
\caption{Swap Local Search}
\label{alg:swap}
\KwIn{Random-key vector $A$}
\KwOut{Best random-key vector $A^{best}$ found in the neighborhood}
Define a vector $RK$ with random order for the random-key indices\;
Update the best solution found $A^{best} \leftarrow A$\;
\For{$i \gets 1$ \textbf{to} $n-1$}{
    \For{$j \gets i+1$ \textbf{to} $n$}{
        Swap random keys $RK_i$ and $RK_j$ of $A$\;
        \If{$\delta_{\mathcal{D}(A)} < \delta_{\mathcal{D}(A^{best})}$}{
            $A^{best} \leftarrow A$\;
        }
        \Else{
            $A \leftarrow A^{best}$\;
        }
    }
}
\Return{$A^{best}$}\;
\end{algorithm}

\subsubsection{Mirror Local Search}
The Mirror Local Search perturbs the random-key values, changing the current value in position $j$ to $(1$ - $A[j])$. Algorithm \ref{alg:invert} illustrates this procedure. Initially, it defines a vector \( RK \) with a random order for the random-key indices and sets the best solution found, \( A^{best} \), to the current solution \( A \). The algorithm then iterates over all indices \( i \) in the random-key vector \( A \). For each index, it inverts the value of the random key at \( RK_i \). After each inversion, if the new solution has a better objective function value than \( A^{best} \), it updates \( A^{best} \) to the new solution. If not, it reverts \( A \) to \( A^{best} \). This process continues until all indices have been processed. Finally, the algorithm returns \( A^{best} \) as the best random-key vector found in the neighborhood.

\begin{algorithm}[h]
\caption{Mirror Local Search}
\label{alg:invert}
\KwIn{Random-key vector $A$}
\KwOut{Best random-key vector $A^{best}$ found in the neighborhood}
Define a vector $RK$ with random order for the random-key indices\;
Update the best solution found $A^{best} \leftarrow A$\;
\For{$i \gets 1$ \textbf{to} $n$}{
    Change the value of the random key $RK_i$ of $A$ to its complement\;
 \If{$\delta_{\mathcal{D}(A)} < \delta_{\mathcal{D}(A^{best})}$}{
        $A^{best} \leftarrow A$\;
    }
    \Else{
        $A \leftarrow A^{best}$\;
    }
}
\Return{$A^{best}$}\;
\end{algorithm}

\subsubsection{Farey Local Search}
The Farey Local Search modifies the value of each random key by randomly selecting values between consecutive terms of the Farey sequence \cite{niven1991introduction}. The Farey sequence of order \(\eta\) consists of all completely reduced fractions between 0 and 1, with denominators less than or equal to \(\eta\), arranged in increasing order. For our purposes, we use the Farey sequence of order 7:
\[
F_7 = \left\{ \frac{0}{1}, \frac{1}{7}, \frac{1}{6}, \frac{1}{5}, \frac{1}{4}, \frac{2}{7}, \frac{1}{3}, \frac{2}{5}, \frac{3}{7}, \frac{1}{2}, \frac{4}{7}, \frac{3}{5}, \frac{2}{3}, \frac{5}{7}, \frac{3}{4}, \frac{4}{5}, \frac{5}{6}, \frac{6}{7}, \frac{1}{1} \right\}
\]
In each iteration of the heuristic, the random keys are processed in a random order. Algorithm \ref{alg:farey} illustrates this procedure. It begins by defining a vector \( RK \) with a random order for the random-key indices and initialises the best solution found, \( A^{best} \), to the current solution \( A \). The algorithm then iterates over each index \( i \) in the random-key vector. For each index \( i \), it iterates over the Farey sequence \( F \) of fractions, setting the value of the random key \( RK_i \) in \( A \) to a random value uniformly generated between \( F_j \) and \( F_{j+1} \), where \( F_j \) and \( F_{j+1} \) are consecutive fractions in the Farey sequence. After updating the random key \( RK_i \), if the new solution has a better objective function value than \( A^{best} \), it updates \( A^{best} \) to the new solution. If not, it reverts \( A \) to \( A^{best} \). The algorithm continues this process until all indices have been processed. Finally, the algorithm returns \( A^{best} \) as the best random-key vector found in the neighborhood.

\begin{algorithm}[h]
\caption{Farey Local Search}
\label{alg:farey}
\KwIn{Random-key vector $A$}
\KwOut{Best random-key vector $A^{best}$ found in the neighborhood}
Define a vector $RK$ with random order for the random-key indices\;
Update the best solution found $A^{best} \leftarrow A$\;
\For{$i \gets 1$ \textbf{to} $n$}{
    \For{$j \gets 1$ \textbf{to} $|F|$}{
        Set the value of the random key $RK_i$ of $A$ with $\textrm{{\fontfamily{pcr}\selectfont UnifRand}}(F_j,F_{j+1})$\;
 \If{$\delta_{\mathcal{D}(A)} < \delta_{\mathcal{D}(A^{best})}$}{
        $A^{best} \leftarrow A$\;
    }
    \Else{
        $A \leftarrow A^{best}$\;
    }
    }
}
\Return{$A^{best}$}\;
\end{algorithm}

\subsubsection{Nelder-Mead Local Search}

\begin{figure}[h]
    \centering
    \includegraphics[width=\textwidth]{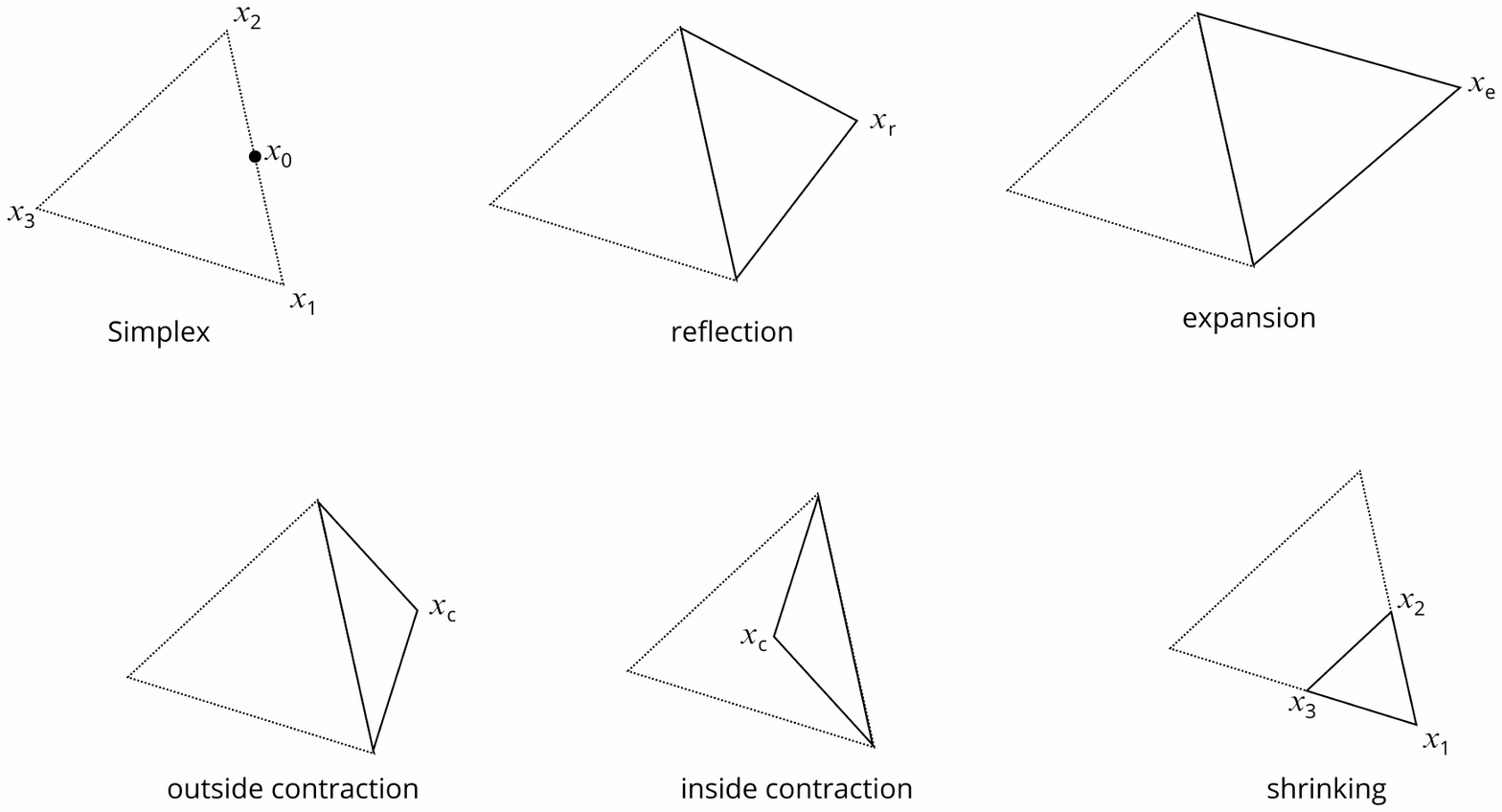} 
    \caption{Illustrative example of the simplex polyhedron and the five moves
of the Nelder-Mead Local Search. Source: \cite{chaves2024randomkeygraspcombinatorialoptimization}.}
    \label{fig:simplex_polyhedron}
\end{figure}

The Nelder–Mead Local Search, introduced by \cite{nelder1965simplex}, is a numerical technique for finding the minimum of an objective function in a multidimensional space. This direct search approach relies on function comparisons and is commonly used in derivative-free nonlinear optimization. The method starts with at least three solutions and can perform five moves: reflection, expansion, inside contraction, outside contraction, and shrinking. In this study, we always apply the Nelder-Mead Local Search with three solutions: \(A_1\), \(A_2\), and \(A_3\), where one is the current solution derived from the metaheuristic, while the others are randomly chosen from a pool of elite solutions found during the search process. These solutions are ordered by objective function value (\(A_1\) is the best and \(A_3\) is the worst). \autoref{fig:simplex_polyhedron} illustrates a simplex polyhedron and the five moves.

Algorithm \ref{NelderMeadAlg} presents the pseudo-code for the Nelder-Mead Local Search adapted for discrete optimization problems. We employ the blending method (Algorithm \ref{alg:blending}) to generate new solutions, using $\rho = 0.5$ and $\mu = 0.02$. The algorithm begins with an initial simplex of three solutions (\(A_1, A_2, A_3\)). The simplex is sorted based on the objective function values, and the simplex’s centroid (\(A_0\)) is computed between \(A_1\) and \(A_2\) (\(A_0 = \text{Blending}(A_1, A_2, 1)\)). The main loop iterates until a termination condition is met. The algorithm performs a series of moves on the simplex during each iteration to explore the search space. A reflection solution (\(A_r = \text{Blending}(A_0, A_3, -1)\)) is computed. If the objective function value at \(A_r\) is better than the current best solution (\(A_1\)), the algorithm computes an expansion solution (\(A_e = \text{Blending}(A_r, A_0, -1)\)). If the objective function value at \(A_e\) is better than at \(A_r\), \(A_3\) is replaced by \(A_e\); otherwise, \(A_3\) is replaced by \(A_r\). If neither the reflection nor the expansion improves the solution, the algorithm contracts towards the solution \(A_r\) or \(A_3\). For an outside contraction (when \(A_r\) is better than \(A_3\)), the contraction solution is \(A_c = \text{Blending}(A_r, A_0, 1)\). For an inside contraction (when \(A_r\) is not better than \(A_3\)), the contraction solution is \(A_c = \text{Blending}(A_0, A_3, 1)\). If the contraction step does not improve, the entire simplex is shrunk towards the best solution \(A_1\) (\(A_i = \text{Blending}(A_1, A_i, 1), i=2,3\)). The algorithm terminates when the maximum number of iterations equals \(n \times e^{-2} \).

\begin{algorithm}
\footnotesize
\caption{Nelder-Mead Local Search. Source: \cite{chaves2024randomkeygraspcombinatorialoptimization}.}\label{NelderMeadAlg}
\KwData{\(A_1, A_2, A_3, n\)}
\KwResult{The best solution found in simplex \(X\)}
\Begin{
    Initialize simplex: \(X \gets \{A_1, A_2, A_3\}\) \;
    Sort simplex \(X\) by objective function value \;
    Compute the simplex centroid \(A_0 \gets \text{Blending}(A_1, A_2, 1)\) \;
    \(iter \gets 0 \) \;
    \(numIter \gets n \cdot \lfloor 1.0/h \rfloor \) \;
    \While{\(iter < numIter\)}{
        \(shrink \gets 0 \) \;
        \(iter \gets iter + 1 \) \;
        Compute reflection solution \(A_r \gets \text{Blending}(A_0, A_3, -1)\) \;
        \If{\(\delta_{\mathscr{D}(A_r)} < \delta_{\mathscr{D}(A_1)}\)}{
            Compute expansion solution \(A_e \gets \text{Blending}(A_r, A_0, -1)\) \;
            \If{\(\delta_{\mathscr{D}(A_e)} < \delta_{\mathscr{D}(A_r)}\)}{
                \(A_3 \gets A_e\) \;
            }
            \Else{
                \(A_3 \gets A_r\) \;
            }
        }
        \Else{
            \If{\(\delta_{\mathscr{D}(A_r)} < \delta_{\mathscr{D}(A_2)}\)}{
                \(A_3 \gets A_r\) \;
            }
            \Else{
                \If{\(\delta_{\mathscr{D}(A_r)} < \delta_{\mathscr{D}(A_3)}\)}{
                    Compute contraction solution \(A_c \gets \text{Blending}(A_r, A_0, 1)\) \;
                    \If{\(\delta_{\mathscr{D}(A_c)} < \delta_{\mathscr{D}(A_r)}\)}{
                        \(A_3 \gets A_c\) \;
                    }
                    \Else{
                        \(shrink \gets 1\) \;
                    }
                }
                \Else{
                    Compute contraction solution \(A_c \gets \text{Blending}(A_0, A_3, 1)\) \;
                    \If{\(\delta_{\mathscr{D}(A_c)} < \delta_{\mathscr{D}(A_3)}\)}{
                        \(A_3 \gets A_c\) \;
                    }
                    \Else{
                        \(shrink \gets 1\) \;
                    }
                }
            }
        }
        \If{\(shrink = 1\)}{
            Replace all solutions except the best \(A_1\) with \(A_i \gets \text{Blending}(A_1, A_i, 1)\), \(i = 2, 3\) \;
        }
        Sort simplex \(X\) by objective function value \;
        Compute the simplex centroid \(A_0 \gets \text{Blending}(A_1, A_2, 1)\) \;
    }
    \Return{\(A_1\)} \;
}
\end{algorithm}

\subsection{Biased Random-Key Genetic Algorithm with Q-Learning} \label{sec:BRKGA-QL}

The BRKGA was introduced by \cite{goncalves2011brkga} and modifies the RKGA by incorporating bias into the parent selection and crossover operators during the evolutionary process. Specifically, BRKGA starts its search process by initializing an initial population denoted as $P$, consisting of randomly generated $|P|$ individuals. In each subsequent generation, these individuals are evaluated based on their fitness, being classified into two distinct groups. The elite population, denoted as $P_e \subset P$, contains the better individuals in terms of fitness. The other group includes non-elite individuals ($P - P_e$).

An individual's fitness is predicated on the resultant value yielded by the objective function \citep{goncalves2011brkga}. In the course of evolution, a new population emerges through a two-step procedure: first, by replicating the elite partition $P_e$ from the current generation, and then, by introducing a novel group of mutants referred to as $P_m$. The quantity of mutants is proportional to the size of the population. The offspring set is denominated as $P_c$ and is produced using the parameterized uniform crossover (PUX) operator \citep{spears1991pux} to complete the new population.

The new population is thus comprised of three distinct parts: $i$) the elite partition $P_e$, $ii$) the mutant partition $P_m$, and $iii$) the offspring set $P_c$ derived from PUX. This operator process hinges on a parameter $\rho_e$, which governs the likelihood of an offspring inheriting a gene (or vector component) from an elite parent instead of a non-elite parent. A random number is generated for each gene within an offspring. Should this random number be less than $\rho_e$, the corresponding gene is inherited from the elite parent's random key; otherwise, it is inherited from the non-elite parent's random key. After this, a decoding mechanism is applied to each individual, mapping the chromosome into a problem solution.

In the version of the BRKGA implemented in this paper, the new population comprises only two components: \( i) \) the elite partition \( P_e \) and \( ii) \) the offspring set \( P_c \). The crossover operator is the blending procedure (see subsection \ref{sec:blending}) with \(factor = 1\) rather than the PUX method. As a result, the role of mutants is replaced by introducing a mutation operator directly within the blending method (lines 2--3 of Algorithm \ref{alg:blending}).

The individuals are then evaluated and sorted based on their fitness. This evolutionary process continues until a predetermined stopping criterion, such as attaining a maximum number of generations, is fulfilled. Additional configurable parameters of the BRKGA encompass the population size ($|P|$), the proportion of individuals contained within the elite partition ($p_e$, where $|P_e| = |P| \times p_e$), and the probability of mutation ($\mu$).

The BRKGA-QL \citep{chaves2021brkgaql} differs from the just presented BRKGA. The first aspect is the replacement of mutants by mutations. Before gene selection from either parent, a mutation may occur with a probability of $\mu$. If positive, the gene acquires a random value within the range of $[0,1)$; otherwise, the gene is inherited based on a probability $\rho_e$, with preference given to the elite parent, and if not, it is inherited from the non-elite parent. Secondly, using a $Q$-Learning agent, it dynamically adjusts the parameter values ($p$, $p_e$, $\mu$, and $\rho_e$). Finally, we have a local search module coupled at the end of each generation. A clustering method is applied to identify promising regions, and an intensification is performed to accelerate the convergence of the method. The RVND (see section \ref{sec:rvnd}) is used as the local search component. Detailed information about this $Q$-learning agent can be found in the following section.

\subsubsection{Q-Learning} \label{sec:gen_frameBRKGA}

The parameter selection process represents an additional optimization challenge, typically addressed through a parameter tuning procedure for later static parameter executions. Such procedure often entails selecting parameter ranges manually or through automation in most existing works \citep{dang2017irace}. In contrast, the $Q$-Learning process offers an automated parameter-tuning mechanism that runs parallel to the main problem-solving algorithm.

Considered one of the most relevant contributions to Reinforcement Learning, the $Q$-Learning algorithm \citep{watkins1992qlearning} is a model-free algorithm (off-policy) that establishes the policy of actions interactively. This algorithm does not require complete modelling of the environment to determine which optimal policy to apply. It is capable of learning from the experiences obtained.

$Q$-Learning has its domain modelled as a Markov Decision Process \citep{sutton1999reinforcement}. It consists of the learning process of the agent inserted in an environment in terms of actions $a$, states $s$, and rewards $r$. Equation \eqref{eq:qLearning} shows the calculations for updating the weights of state-action pairs. Function $Q$ delineates the value associated with the state-action pair $(s, a)$ at the $Q$-table $i+1$, symbolising the quality of actions taken to anticipate future returns or rewards. This update is contingent on two parameters: the learning factor, denoted as $\lambda^{lf}$, and the diversification factor, denoted as $\lambda^{df}$.

\begin{flalign}
    Q^{i+1}(s,a) := Q^{i}(s,a) + \lambda^{lf} [R^{i}(s,a) + \lambda^{df} \times max_{a'} Q^{i}(s', a') - Q^{i}(s,a)]  \label{eq:qLearning}
\end{flalign}

Moreover, analogous to metaheuristic common practices, the $Q$-Learning execution tends to take the best-known actions for each state (intensification) to obtain the expected maximum return. On the other hand, it is also necessary to choose different actions to explore other policies (diversification). The $\epsilon$-Greedy policy consists of a good strategy to balance intensification and diversification, choosing the action with the highest value in $Q$ with probability $1-\epsilon$ or selecting an action at random with probability $\epsilon$. The $Q$-table represents the learning process with the value of each state-action pair. 

In our implementation, at the end of each BRKGA-QL generation $j$, a reward $R^{i}(s,a)$ is incremented for the corresponding state-action pair $(s, a)$ and $Q$-table $i$, as showed by Equation \eqref{eq:reward}, where $\delta_{b_{j}}$ is the best fitness in the current generation and $\delta_{b_{j-1}}$ the previous one.

\begin{equation} \label{eq:reward}
 R^{i}(s,a) \gets R^{i}(s,a) + \left\{\begin{matrix}
 \frac{(\delta_{b_{j-1}}/\delta_{b_{j}})-1}{|P|}, & \mbox{if  } \delta_{b_{j}} < \delta_{b_{j-1}}
 \\ 
 0, & otherwise
\end{matrix}\right.
\end{equation}

Furthermore, the reward function serves as an intermediary solution, striking a balance between enhancing the current best fitness and the binary rewards as proposed by \cite{karafotias2015evaluating}. Notably, this function exhibits a higher sensitivity to scale than the binary reward, considering the magnitude of improvement and the population size. \autoref{tab:parameterStates} presents the potential states for each parameter, with  $\lambda^{df}$ fixed as $0.8$ and $\lambda^{lf}$ starts with $1$ and decreases linearly until $0.1$ at the time limit. The $Q$-table initially has all values set to $0$. It undergoes updates every $k$ BRKGA-QL generation using Equation \eqref{eq:qLearning}, followed by the reset of corresponding rewards.

\begin{table}[htbp]
\centering
\caption{Possible parameter states.}
\label{tab:parameterStates}
\scalebox{0.75}{
\begin{tabular}{l|r} 
Parameter & States \\ \hline
$|P|$ & 233, 377, 610, 987, 1597, 2584 \\
$p_e$ & 0.10, 0.15, 0.20, 0.25, 0.30 \\
$\mu$ & 0.01, 0.02, 0.03, 0.04, 0.05 \\
$\rho_e$ & 0.55, 0.60, 0.65, 0.70, 0.75, 0.80
\end{tabular}}
\end{table}

Finally, Algorithm \ref{alg:brkga-ql} presents the summarised implementation of the BRKGA-QL. It is comprised of a usual BRKGA implementation \citep{goncalves2011brkga} coupled with a local search (Lines \ref{algql:lsBeg}$-$\ref{algql:lsEnd}) and the added QL component (Lines \ref{algql:ql1}, \ref{algql:ql2}, \ref{algql:ql3} and \ref{algql:ql4}). The point of integration between the problem at hand (IORSP) and BRKGA implementations, including the present one, lies in the decoder function. This function receives an individual (a vector of random keys) and deterministically maps it into a feasible solution, a process that is elaborated upon in the section \ref{sec:decoder}.

\begin{algorithm}[H]
\caption{BRKGA-QL}
\label{alg:brkga-ql}
\KwIn{Time limit $TL$}
\KwOut{Best solution found $A^{best}$}
\textbf{Step 1:} Initialize $Q$-Table values\; \label{algql:ql1}
\textbf{Step 2:} Randomly generate the population $P$\;
\textbf{Step 3:} Evaluate and sort $P$ by fitness. Store the best individual in $A^{best}$\;

\While{$TL$ is not reached}{
    \textbf{Step 4:} Set $Q$-Learning parameters ($\epsilon, lf, df$)\; \label{algql:ql2}
    \textbf{Step 5:} Choose an action for each parameter ($p, p_e, \mu, \rho_e$) from the $Q$-Table using the $\epsilon$-greedy policy\; \label{algql:ql3}
    \textbf{Step 6:} Evolutionary process\;
    Classify $P$ as elite or non-elite individuals\;
    Create elite set $P_e$ using $p_e$ as a guide\;
    Create the offspring set $P_c$ through the blending procedure, using $\rho_e$, $\mu$, and $factor=1$ as guides\;
    $P \gets P_e \cup P_c$\;
    Evaluate and sort $P$ by fitness\;
    \If{the best individual improved}{
        Store the best individual in $A^{best}$\;
        \textbf{Step 7:} Set reward ($R^i$) and update $Q$-Table\; \label{algql:ql4}
    }
    \If{exploration or stagnation is detected}{ \label{algql:lsBeg}
        \textbf{Step 8:} Local Search\;
        Identify communities in $P_e$ with the clustering method\;
        Apply RVND in the best individuals of these communities\;  \label{algql:lsEnd}
        Apply Shaking in other individuals\;
    }
}
\Return $A^{best}$\;
\end{algorithm}

\subsection{Simulated Annealing}\label{secSA}

Simulated Annealing (SA) is a widely used global optimization method inspired by statistical physics \cite{Kirkpatrick1983, Cerny1985} and is supported by theoretical guarantees of global convergence \cite{Dekkers1991}. SA starts with an initial solution (denoted as a vector of random keys) $A\in S$, where $S$ represents the random-key solution space. Then, a neighbourhood solution $A^{'}$ is generated through a perturbation algorithm. The search procedure in SA is based on the Metropolis acceptance criterion of \cite{metropolis1953}, which models how a thermodynamic system moves from the current solution (state) to a candidate solution in which the objective function (energy content) is being minimised. If $\delta_{\mathscr{D}(A^{'})}-\delta_{\mathscr{D}(A)}\leq 0$, then $A^{'}$ is accepted as the current solution, else the candidate solution is accepted based on the acceptance probability:
\begin{equation}\label{SA1}
    P(A^{'})=\exp\left(-\frac{\Delta E}{T}\right),
\end{equation}
where $\Delta E= \delta_{\mathscr{D}(A^{'})}-\delta_{\mathscr{D}(A)}$ and $T$ defines the current temperature. The key idea is to prevent the algorithm from becoming trapped in local optima by allowing uphill moves. It is important to note that uphill moves are more likely to occur at higher temperatures.

Algorithm \ref{SAalg} presents the SA pseudo-code used in this work. The procedure starts by initializing with a solution represented by a random-key vector $A$. The algorithm runs a loop until a stopping criterion is met. Within this loop, it generates a neighboring solution ($A^{'}$) from the current solution ($A$) and calculates the difference in objective function values (energy difference, \(\Delta E\)). If the new solution has a lower energy (\(\Delta E \leq 0\)), it is accepted as the current solution and potentially updates the best-found solution. If the new solution has higher energy (\(\Delta E > 0\)), it may still be accepted based on a probability determined by the Metropolis criterion via Equation \eqref{SA1}, which allows for occasional uphill moves to escape local optima. The temperature ($T$) is then updated by multiplying it with a cooling rate (\(\alpha\)). An RNVD is subsequently called to refine the current solution further. This process continues, gradually reducing the temperature and thus the probability of accepting worse solutions until the stopping criterion is satisfied. The algorithm finally returns the best solution found during the search.

\begin{algorithm}[H]
\caption{Simulated Annealing}\label{SAalg}
\KwData{Random-key vector $A$, initial temperature $T_0$, cooling rate $\alpha$, $\beta_{min}$, $\beta_{max}$, $SA_{max}$, time limit $TL$}
\KwResult{Best found solution}
$A^{best} \gets A$, $T \leftarrow T_0$ \; 
\While{$TL$ is not reached}{
    $iter \leftarrow 0$\;
    \While{$iter < SA_{max}$}{
        $A' \gets \text{Shaking}(A, \beta_{min}$, $\beta_{max})$\;
        Calculate the energy difference $\Delta E \gets \delta_{\mathscr{D}(A^{'})}-\delta_{\mathscr{D}(A)}$\;
        \If{$\Delta E \leq 0$}{
            $A \gets A'$\;
            \If{$\delta_{\mathscr{D}(A)} < \delta_{\mathscr{D}(A^{best})}$}{
                $A^{best} \gets A$\;
            }
        }
        \Else{
            Calculate the acceptance probability $P \gets \exp\left(-\frac{\Delta E}{T}\right)$\;
            Generate a random number $r \in [0, 1]$\;
            \If{$r < P$}{
                $A \gets A'$\;
            }
        }
        $iter++$\;
    }
    Update temperature $T \gets \alpha \times T$\;
    $A \gets \text{RVND}(A)$\;
}
\Return{$A^{best}$}
\end{algorithm}

\subsection{Iterated Local Search}\label{secILS}

The Iterated Local Search (ILS) is a straightforward yet powerful heuristic to solve various optimization problems. The foundational idea of ILS was introduced by \cite{baxter1981local} and further elaborated by \cite{lourenco2003iterated}. The essence of ILS lies in iteratively constructing a sequence of solutions using a specific heuristic, which typically leads to significantly better solutions than repeated random trials of the same heuristic. This optimization technique attempts to escape local optima by applying local search and perturbation in an iterative manner.


The ILS algorithm, as detailed in Algorithm \ref{ISLalg}, begins with a random-key vector $A$ as input. Initially, the vector $A$ undergoes an RVND process, resulting in the solution set as $A^{best}$. The algorithm then enters a loop until the stopping criterion is met. A copy of $A^{best}$, denoted as $A'$, is created and perturbed within the loop. This perturbed solution is then refined using the RVND function. If the objective function value $\delta_{\mathscr{D}(A')}$ of the newly obtained solution $A'$ is better than that of the current best solution $A^{best}$, represented as $\delta_{\mathscr{D}(A^{best})}$, then $A^{best}$ is updated to $A'$. Once the stopping criterion is satisfied, the algorithm returns $A^{best}$ as the best-found solution.

\begin{algorithm}[H]
\caption{Iterated Local Search}\label{ISLalg}
\KwData{Random-key vector $A$, $\beta_{min}$, $\beta_{max}$, time limit $TL$}
\KwResult{Best found solution}

$A \gets \text{RVND}(A)$\;
$A^{best} \gets A$\;

\While{$TL$ is not reached}{
    $A' \gets A^{best}$\;
    $A' \gets \text{Shaking}(A', \beta_{min}$, $\beta_{max})$\;
    $A' \gets \text{RVND}(A')$\;
    
    \If{$\delta_{\mathscr{D}(A^{'})} < \delta_{\mathscr{D}(A^{best})}$}{
        $A^{best} \gets A'$\;
    }
   
}
\Return{$A^{best}$}\;

\SetAlgoNlRelativeSize{-1}
\end{algorithm}

\subsection{Encoding and Decoding} \label{sec:decoder}

Solutions to the IORSP are encoded with a vector $A$ of $n=|K|$ random keys, where $K$ is the set of surgeries. Then, each random key corresponds to a unique surgery. Our decoder implementation follows a straightforward approach: $i$) sorts the surgeries based on their random-key values, $ii$) sorts the surgeries based on their priority values, keeping the random-key order for equal priorities; $iii$) schedules each surgery task sequentially at the earliest available time slots.

\autoref{tab:dynamic} shows the decoder's dynamic structures and operators. We have a solution problem $s_A$ that contains, for each surgery $k \in K$, a surgery allocation data $D_k$, and each surgery allocation data contains an allocated room $r_k$, set of equipment $e_t$, starting time $\sigma_t$ and blocking time $\psi_t$. Thus, the scheduled surgeries makespan $\delta_{\mathscr{D}(A)}$ is when the last patient completes its last task.

\begin{table}[h]
\centering
\caption{Decoder Notations}
\label{tab:dynamic}
\scalebox{0.75}{
\begin{tabular}{l|l} \hline 
\textbf{Solution variables}     &      \\ 
$s_A$        & Solution problem construct from random-key vector $A$      \\
$D_k$        & Surgery's $k$ allocation data \\
$r_t$        & Room for task $t$  \\
$e_t$        & Set of equipment for task $t$ \\
$\sigma_t$   & Starting time for task $t$  \\ 
$\psi_t$     & Blocking time for task $t$  \\  
$\delta_{\mathscr{D}(A)}$   & Makespan of random-key vector $A$      \\ \hline
\textbf{Operators}    &     \\
$Next(t)$   &  Returns the next task to be executed after $t$ on the same surgery\\
$First(k)$    &  Returns the first task to be executed on surgery $k$ \\
$Last(k)$    &  Returns the last task to be executed on surgery $k$ \\
$Search(R_t, k, s_A, \sigma^{min})$  & Returns the earliest available resource $r$ and the corresponding candidate \\
              &  moment $\sigma^{c}_{r}$ at which task $k$ can be executed in solution $s_A$, after moment $\sigma^{min}$. \\  \hline 
\end{tabular}}
\end{table}

Algorithm \ref{alg:greedDecoder} details our implementation. The solution starts with no surgeries scheduled (Line \ref{alg:dec:emptySol}) and, consequently, $\delta_{\mathscr{D}(A)}=0$, then the surgeries are sorted given the sequence of random keys (Line \ref{alg:dec:sortRk}). Then, given the final sequence of surgeries, each one is scheduled by our Greed Insertion algorithm (Line \ref{alg:dec:useGI}).


\begin{algorithm}[!ht]
    \footnotesize
    \KwData{Vector $A$ of random keys}
    \KwResult{Fitness (makespan) value, $\delta_{\mathscr{D}(A)}$}
    $s_A \gets \emptyset$; $\delta_{\mathscr{D}(A)} \gets 0$\; \label{alg:dec:emptySol}
    $SortByRK(A)$\; \label{alg:dec:sortRk}
    \For{$a \in A$}{ \label{alg:dec:mlStart}
       $GreedInsertion(s_A, K_{a}, \delta_{\mathscr{D}(A)})$\; \label{alg:dec:useGI}
    } \label{alg:dec:mlEnd}
    \Return $\delta_{\mathscr{D}(A)}$ \label{alg:dec:End}
    \caption{Decoder}
    \label{alg:greedDecoder}
\end{algorithm}

Algorithm \ref{alg:greedInsertion} details the Greed Insertion algorithm. It also has the simple concept of scheduling each surgery's task as soon as possible, given all available rooms, types of equipment, and persons. It receives a current problem solution $s_A$ and a surgery $k$ to be scheduled and starts by setting variable $\sigma^m$ (minimal start time) to zero and $Success$ to false (Line \ref{alg:gi:start}). The main loop (Lines \ref{alg:gi:fLoopS}$-$\ref{alg:gi:fLoopE}), for each task (Line \ref{alg:gi:sLoopS}), searches all compatible rooms $R_t$ and selects it the one $r_t$ with the earliest moment available $\sigma_c$ (Line \ref{alg:gi:searchRoom}), with the assistance of operator $Search$ as noted on Table \ref{tab:dynamic}. The same logic is applied to select the required equipment types $F_t$, if any is required (Lines \ref{alg:gi:startSE}$-$\ref{alg:gi:endSE}) and for the required people $P_t$ (Lines \ref{alg:gi:startSP}$-$\ref{alg:gi:endSP}).

Variable $\sigma_t$ indicates the selected start time for task $k$. It starts as $\sigma^{min}$ (Line \ref{alg:gi:startTime}), and it is updated with the maximum value of the selected room (Line \ref{alg:gi:startTimeR}), each selected equipment (Line \ref{alg:gi:startTimeE}) and each involved person's schedule (Line \ref{alg:gi:startTimeP}). The blocking value $\phi_t$ is computed as the delay (Line \ref{alg:gi:block}), if it exceeds the limit (Line \ref{alg:gi:blockCond}), the new $\sigma^{min}$ is set by delaying the first task (Line \ref{alg:gi:delay}) and restarting $k$'s scheduling (Line \ref{alg:gi:restart}), otherwise $\sigma^{min}$ is updated to allocate the following task (Line \ref{alg:gi:nextT}). After a successful surgery allocation, the surgery allocation data $D_k$ is added to solution $s_A$ (Line \ref{alg:gi:insertSD}) and hence, the makespan $\delta_{\mathscr{D}(A)}$ is updated (Line \ref{alg:gi:updateMakespan}).

\begin{algorithm}[!ht]
    \footnotesize
    \KwData{Solution $s_A$, surgery $k$, and the current makespan $\delta_{\mathscr{D}(A)}$}
    \KwResult{A Solution $s_A$ with surgery $k$ scheduled.}
    $\sigma^{m} \gets 0; Success \gets False$\; \label{alg:gi:start}
    \While{$\neg Success$}{ \label{alg:gi:fLoopS}
        $Success \gets True$; $D_k \gets \emptyset$\; \label{alg:gi:sucess}
        \For{Task $t \in T_k$}{ \label{alg:gi:sLoopS}
            $\sigma_t \gets \sigma^{min}$\; \label{alg:gi:startTime}
            $(r_{t}, \sigma^{c}) \gets Search( R_{t}, t, s_A, \sigma_t )$\; \label{alg:gi:searchRoom}
            $\sigma_t \gets max(\sigma_t, \sigma^{c})$\; \label{alg:gi:startTimeR}
            $e_t \gets \emptyset $\; \label{alg:gi:emptyEqpSet}
            \For{$f \in F_{t}$}{ \label{alg:gi:startSE}
                $(e, \sigma^{c}) \gets Search( E_{f}, t, s_A, \sigma_t )$\;
                $e_t \gets e_t \bigcup \{ e \} $\;
                $\sigma_t \gets max(\sigma_t, \sigma^{c})$\; \label{alg:gi:startTimeE}
            } \label{alg:gi:endSE}
            \For{$p \in P_{t}$}{ \label{alg:gi:startSP}
                $(p, \sigma^{c}) \gets Search( \{ p \}, t, s_A, \sigma_t )$\;
                $\sigma_t \gets max(\sigma_t, \sigma^{c})$\; \label{alg:gi:startTimeP}
            } \label{alg:gi:endSP}
            $\phi_t \gets \sigma_t - \sigma^{min}$\; \label{alg:gi:block}
            \eIf{$\phi_t > \Phi$}{ \label{alg:gi:blockCond}
                $\sigma^{min} \gets \sigma_{First(k)} + \phi_t$\; \label{alg:gi:delay}
                $Success \gets False$; $Break$\; \label{alg:gi:restart}
            }{
                $\sigma^{m} \gets \sigma_t + \gamma_{t}^{d} + \gamma_{t}^{m}$\; \label{alg:gi:nextT}
            }
            $D_k \gets D_k \bigcup \{ (r_{t}, e_{t}, \sigma_{t}, \phi_{t} \}$\; \label{alg:gi:add2ID}
        } \label{alg:gi:sLoopE}
    } \label{alg:gi:fLoopE}
    $s_A \gets s_A \bigcup \{ D_k \}$\; \label{alg:gi:insertSD}
    $\delta_{\mathscr{D}(A)} \gets max(\delta_{\mathscr{D}(A)}, \sigma_{Last(k)} + \gamma_{Last(k)}^{d})$\; \label{alg:gi:updateMakespan}
    \caption{Greed Insertion}
    \label{alg:greedInsertion}
\end{algorithm}

\begin{figure}[!ht]
    \centering
    \scalebox{0.25}{\includegraphics{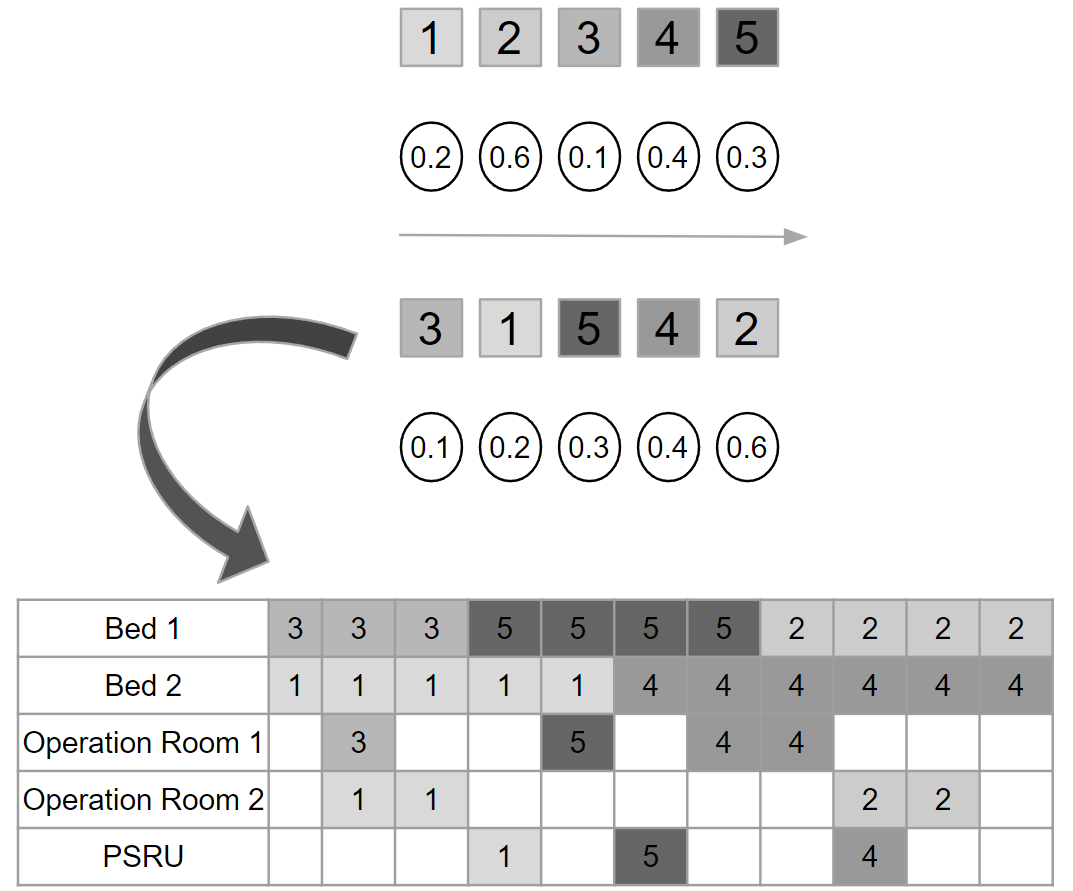}}
    \caption{Example of our decoder with a Greed Insertion strategy.}
    \label{fig:decoder}
\end{figure}

\autoref{fig:decoder} illustrates an example of the Algorithm \ref{alg:greedDecoder} on a simplified scenario with five surgeries, no priorities, two beds, two ORs, one PSRU, and all subjects with clear initial schedules. Notably, Surgeries $2$ and $3$ lack PSRU tasks; not all tasks follow the same room sequencing, and all surgeries allocate first and last the same bed as described in Section \ref{sec:intro}.

The decoder receives a vector of random keys $A=\{0.2,0.6,0.1,0.4,0.3\}$. Subsequently, each surgery is scheduled in the sequence $3$, $1$, $5$, $4$, and $2$. Firstly, Surgery $3$ involves $3$ tasks, with the first task allocating Bed 1 at the first time slot, OR 1 at the second time slot, and the patient returning to Bed 1 at the third time slot. Surgery $1$ comprises $4$ tasks, with the first allocating Bed 1 at the first time slot, OR 2 at the second and third time slots, the PSRU at the fourth, and the patient returning to Bed 2 at the fifth. Similarly, Surgeries $5$, $4$, and $2$ also have their respective tasks sequentially allocated at the first available moment.

\section{Computational experiments and analysis} \label{sec:compExp}

Our proposed metaheuristics were coded in C++ and compiled with GCC. The proposed lower bound models were coded on Python and solved with Gurobi $10.01$ \citep{gurobi}. The computer used in all experiments was a Dual Xenon Silver 4114 20c/40t 2.2Ghz processor with 96GB of DDR4 RAM and running CentOS 8.0 x64. Each proposed model was solved for each instance with a time limit of $3600$ seconds, and the metaheuristics were run five times for each instance with a time limit proportional to the number of surgeries. The literature instances and results were retrieved from \cite{burdett2018integFJSS}, and our case study instances were generated based on gathered data from multiple sources. This process is detailed in Section \ref{sec:results-caseStudy}. \autoref{tab:instanceData} presents the general characteristics of each instance, with its name, number of surgeries and number of rooms per room type in the sequence a patient usually has, and the total CPU time limit (TL) of each heuristics run.

The parameters of the SA and ILS are tuned using an offline strategy. We applied an experimental design approach considering potential values for each parameter and identifying the ``best'' value for each parameter through many experiments on a subset of the problem instances. The parameters of the ILS were $\beta_{min}=0.10$ and $\beta_{max}=0.20$. The parameters of the SA were $T_0=1000000$, $\alpha=0.99$, $SA_{max} = 200$ $\beta_{min}=0.05$, and $\beta_{max}=0.20$. The parameters of the BRKGA-QL are tuned during the search process using the $Q$-Learning method.


\begin{table}[htbp]
\centering
\caption{Literature and case study instances details.}
\label{tab:instanceData}
\scalebox{0.70}{
\begin{tabular}{lccr|lccr} \hline
\multicolumn{4}{c}{Literature}                     & \multicolumn{4}{|c}{Case Study}                   \\ \hline
Name     & \#Surgeries & \#Rooms     & TL(s) & Name & \#Surgeries & \#Rooms     & TL(s) \\ \hline
CASE\_01.dat & 50        & 5,2,5,10  & 120         & p070     & 58        & 31,3,2,4  & 60        \\
CASE\_02.dat & 100       & 5,2,5,10  & 200         & p078     & 66        & 23,3,3,4  & 70        \\
CASE\_03.dat & 150       & 5,2,5,10  & 240         & p087     & 75        & 29,3,3,4  & 80        \\
CASE\_04.dat & 200       & 5,2,5,10  & 300         & p093     & 81        & 30,3,3,4  & 90        \\
CASE\_05.dat & 50        & 5,3,5,10  & 120         & p098     & 86        & 40,3,2,4  & 100       \\
CASE\_06.dat & 100       & 5,3,5,10  & 200         & p100     & 84        & 39,4,3,5  & 100       \\
CASE\_07.dat & 150       & 5,3,5,10  & 240         & p109     & 93        & 35,4,3,5  & 100       \\
CASE\_08.dat & 200       & 5,3,5,10  & 300         & p120     & 104       & 46,4,4,5  & 100       \\
CASE\_09.dat & 50        & 5,4,5,20  & 120         & p138     & 118       & 43,5,4,5  & 120       \\
CASE\_10.dat & 100       & 5,4,5,20  & 200         & p153     & 133       & 71,5,4,7  & 120       \\
CASE\_11.dat & 150       & 5,4,5,20  & 240         & p165     & 141       & 78,6,4,7  & 130       \\
CASE\_12.dat & 200       & 5,4,5,20  & 300         & p178     & 154       & 88,6,4,6  & 130       \\
CASE\_13.dat & 50        & 5,5,5,20  & 120         & p183     & 159       & 89,6,4,6  & 140       \\
CASE\_14.dat & 100       & 5,5,5,20  & 200         & p189     & 161       & 71,7,5,7  & 140       \\
CASE\_15.dat & 150       & 5,5,5,20  & 240         & p192     & 164       & 72,7,4,8  & 140       \\
CASE\_16.dat & 200       & 5,5,5,20  & 300         & p193     & 165       & 78,7,5,7  & 140       \\
CASE\_17.dat & 50        & 5,10,5,20 & 120         & p197     & 173       & 99,6,5,7  & 160       \\
CASE\_18.dat & 100       & 5,10,5,20 & 200         & p201     & 177       & 73,6,4,8  & 160       \\
CASE\_19.dat & 150       & 5,10,5,20 & 240         & p216     & 188       & 95,7,5,9  & 160       \\
CASE\_20.dat & 200       & 5,10,5,20 & 300         & p233     & 197       & 88,9,4,10 & 160       \\
CASE\_21.dat & 50        & 5,10,5,30 & 120         &          &           &           &             \\
CASE\_22.dat & 100       & 5,10,5,30 & 200         &          &           &           &             \\
CASE\_23.dat & 150       & 5,10,5,30 & 240         &          &           &           &             \\
CASE\_24.dat & 200       & 5,10,5,30 & 300         &          &           &           &  \\ \hline         
\end{tabular}}
\end{table}


We analyzed the computational results in terms of quality and robustness. We present the average CPU time and the relative percentile deviation (RPD) for the best solution found ($X_{min}$) as shown in Equation \ref{eq:rpd_min}. The RPD values are relative deviations from reference ones. We calculated the RPD in each run and presented the average and best values. A statistical analysis is made per table between the resulting average RPDs. First, we classify it as parametric or non-parametric and then compare it with the Paired T Student Test group, if parametric, or with the Wilcoxon test otherwise. The algorithms are compared in pairs with the null hypothesis first; if rejected, the ``less'' alternative hypothesis is tested, and superior performance is considered given statistical confidence greater than $95\%$ ($p$-value $\leq 0.05$). 

\begin{flalign}
    RPD = (X/X_{min} - 1) \times 100 \label{eq:rpd_min} 
\end{flalign}



\subsection{Computational results for literature instances}\label{sec:results-literature}

The case study modelling of \cite{burdett2018integFJSS} does not completely match ours. The main difference is that, in their case, the wardroom is the long-time recovery room; in our case, the patient returns to the initial bed for a long recovery. In both cases, one type of room is reserved during each patient's whole stay. This allowed their case to be compatible with Formulation \eqref{mod1:FO}-\eqref{mod1:rest4}. For these instances, the distinction between the scenarios with and without business hours lies in the fact that, in the case with it, the ORs are only operational during regular business hours (from 8:00 AM to 5:00 PM from Monday until Friday), and the other rooms are always available, these are implemented with availability slots. Other distinctions include the absence of surgeon or equipment scheduling, compatibility between rooms and surgery types, and customisation in initial schedules. These variances were more straightforward to accommodate when we examined a simplified scenario based on our case study. In the following tables, the displayed lower bounds are the best ones computed by any of our relaxed formulations; the formulations and their results are detailed on \ref{sec:relaxed_model_bed}$-$\ref{sec:relaxed_model_results}.

\autoref{tab:resultLitHeuristicNoTW} shows our heuristic results with the literature instances without business hours. The old lower bound and Hybrid Simulated Annealing (HSA) results are reported from \cite{burdett2018integFJSS}, algorithms were coded on C++ and ran on a personal computer with a $2.6$Ghz processor and 16GB of RAM under Windows 7. The first tree columns describe the instances, followed by the literature lower bound (Old), our best computed lower bound from \autoref{tab:resultLiteratureForm} (New), followed by the HSA and RKO heuristic results, the makespan values for best solution found (best) in days, its best and average relative percentile deviation, respectively BRPD and ARPD, and the average time to the best solution (ATTB) in seconds. Analysing the results, our relaxed formulations computed lower bound values better than the ones in the literature. Besides, the RKO methods found better results for all literature instances without business hours. The SA found seven new best-known solutions (BKS), the ILS found 15 new BKS, and the BRKGA-QL found four new BKS. These methods are robust with very small RPDs. The statistical tests concluded that all RKO heuristics performed better than the literature heuristic, and between the RKO heuristics, the ILS was superior to the other two ($p$-values return by the Wilcoxon test were 7.35E-11 between ILS and BRKGA-QL and 0.0013 between ILS and SA), and the SA was superior to the BRKGA-QL ($p$-value was 4.13E-08). The ATTB shows us that the BRKGA-QL was the fastest to converge to the best solution found. Looking at the obtained new lower bounds and upper bounds, instance ``CASE\_21'' has an optimal result proved, and such a solution was found by all three RKO heuristics.

\begin{table}[htbp]
\renewcommand{\arraystretch}{1.1}
\centering
\caption{Literature instances heuristic results without business hours.}
\label{tab:resultLitHeuristicNoTW}
\resizebox{\textwidth}{!}{%
\begin{tabular}{l|rr|rrr|rrrr|rrrr|rrrr} \hline
 & \multicolumn{2}{c|}{Lower Bounds} & \multicolumn{3}{c|}{HSA} & \multicolumn{4}{c|}{SA} & \multicolumn{4}{c|}{ILS} & \multicolumn{4}{c}{BRKGA-QL} \\ \hline
\multicolumn{1}{c|}{Instance} & \multicolumn{1}{c}{Old} & \multicolumn{1}{c|}{New} & \multicolumn{1}{c}{best} & \multicolumn{1}{c}{BRPD(\%)} & \multicolumn{1}{c|}{time (s)} & \multicolumn{1}{c}{best} & \multicolumn{1}{c}{BRPD(\%)} & \multicolumn{1}{c}{ARPD(\%)} & \multicolumn{1}{c|}{ATTB(s)} & \multicolumn{1}{c}{best} & \multicolumn{1}{c}{BRPD(\%)} & \multicolumn{1}{c}{ARPD(\%)} & \multicolumn{1}{c|}{ATTB(s)} & \multicolumn{1}{c}{best} & \multicolumn{1}{c}{BRPD(\%)} & \multicolumn{1}{c}{ARPD(\%)} & \multicolumn{1}{c}{ATTB(s)} \\ \hline
CASE\_01 & 14.82 & 16.41 & 16.80 & 1.54 & 172.8 & 16.55 & \textbf{0.00} & 0.04 & 83.37 & 16.55 & 0.04 & 0.07 & 60.03 & 16.56 & 0.07 & 0.11 & 33.76 \\
CASE\_02 & 32.24 & 35.32 & 36.04 & 1.73 & 1821.6 & 35.43 & 0.02 & 0.04 & 136.49 & 35.43 & \textbf{0.00} & 0.02 & 145.74 & 35.44 & 0.05 & 0.07 & 20.56 \\
CASE\_03 & 44.29 & 49.06 & 50.10 & 1.86 & 6282 & 49.19 & 0.004 & 0.01 & 206.45 & 49.19 & \textbf{0.00} & 0.01 & 149.66 & 49.19 & 0.004 & 0.02 & 43.57 \\
CASE\_04 & 60.81 & 67.05 & 68.13 & 1.43 & 16110 & 67.17 & 0.001 & 0.02 & 247.83 & 67.17 & \textbf{0.00} & 0.01 & 240.66 & 67.18 & 0.02 & 0.04 & 101.91 \\
CASE\_05 & 15.37 & 16.88 & 17.41 & 2.70 & 151.2 & 16.95 & \textbf{0.00} & 0.02 & 67.31 & 16.95 & 0.004 & 0.03 & 73.97 & 16.96 & 0.07 & 0.11 & 55.08 \\
CASE\_06 & 29.89 & 32.94 & 34.49 & 4.47 & 1605.6 & 33.02 & 0.01 & 0.02 & 89.23 & 33.01 & \textbf{0.00} & 0.02 & 116.35 & 33.03 & 0.04 & 0.08 & 53.68 \\
CASE\_07 & 44.18 & 48.81 & 49.45 & 1.17 & 6080.4 & 48.88 & 0.01 & 0.03 & 147.67 & 48.88 & \textbf{0.00} & 0.01 & 172.58 & 48.89 & 0.02 & 0.04 & 71.14 \\
CASE\_08 & 61.27 & 67.50 & 68.73 & 1.72 & 11998.8 & 67.57 & \textbf{0.00} & 0.01 & 161.91 & 67.57 & 0.002 & 0.01 & 208.13 & 67.58 & 0.01 & 0.03 & 93.41 \\
CASE\_09 & 6.79 & 7.60 & 8.17 & 5.82 & 122.4 & 7.73 & 0.11 & 0.14 & 87.15 & 7.73 & 0.13 & 0.20 & 113.37 & 7.72 & \textbf{0.00} & 0.33 & 42.76 \\
CASE\_10 & 14.99 & 16.53 & 17.54 & 5.13 & 1396.8 & 16.69 & 0.02 & 0.07 & 146.13 & 16.68 & \textbf{0.00} & 0.06 & 110.71 & 16.69 & 0.06 & 0.15 & 126.59 \\
CASE\_11 & 22.09 & 24.31 & 25.95 & 6.11 & 5230.8 & 24.46 & 0.03 & 0.06 & 189.41 & 24.46 & \textbf{0.00} & 0.04 & 140.09 & 24.47 & 0.08 & 0.1 & 56.42 \\
CASE\_12 & 30.57 & 33.60 & 35.13 & 4.12 & 12834 & 33.74 & \textbf{0.00} & 0.05 & 207.66 & 33.75 & 0.02 & 0.06 & 251.78 & 33.75 & 0.04 & 0.09 & 114.96 \\
CASE\_13 & 7.34 & 8.11 & 9.08 & 10.47 & 118.8 & 8.23 & 0.11 & 0.20 & 76.66 & 8.22 & \textbf{0.00} & 0.17 & 53.77 & 8.23 & 0.14 & 0.26 & 52.03 \\
CASE\_14 & 15.41 & 16.92 & 18.72 & 9.76 & 1317.6 & 17.06 & 0.04 & 0.06 & 137.33 & 17.06 & \textbf{0.00} & 0.02 & 114.33 & 17.06 & 0.03 & 0.14 & 48.61 \\
CASE\_15 & 23.31 & 25.50 & 27.11 & 5.82 & 4788 & 25.63 & 0.02 & 0.07 & 152.14 & 25.62 & \textbf{0.00} & 0.05 & 120.20 & 25.64 & 0.09 & 0.15 & 129.74 \\
CASE\_16 & 32.23 & 35.27 & 37.02 & 4.58 & 12394.8 & 35.41 & 0.03 & 0.04 & 226.79 & 35.4 & 0.004 & 0.04 & 289.84 & 35.40 & \textbf{0.00} & 0.04 & 188.69 \\
CASE\_17 & 8.23 & 9.03 & 9.79 & 7.66 & 97.2 & 9.09 & \textbf{0.00} & 0.16 & 82.16 & 9.12 & 0.30 & 0.31 & 66.70 & 9.12 & 0.30 & 0.41 & 81.80 \\
CASE\_18 & 15.46 & 16.99 & 18.22 & 6.74 & 1202.4 & 17.08 & 0.03 & 0.07 & 164.61 & 17.07 & \textbf{0.00} & 0.06 & 116.01 & 17.07 & 0.02 & 0.13 & 52.89 \\
CASE\_19 & 23.91 & 26.23 & 27.67 & 5.17 & 4449.6 & 26.32 & 0.04 & 0.09 & 157.74 & 26.31 & \textbf{0.00} & 0.05 & 226.69 & 26.32 & 0.03 & 0.1 & 87.91 \\
CASE\_20 & 30.48 & 33.55 & 35.29 & 4.95 & 11134.8 & 33.65 & 0.06 & 0.08 & 284.51 & 33.63 & \textbf{0.00} & 0.03 & 212.56 & 33.63 & 0.01 & 0.08 & 135.85 \\
CASE\_21 & 5.04 & \textbf{6.09} & 6.54 & 7.36 & 118.8 & 6.09 & \textbf{0.00} & 0.00 & 45.72 & 6.09 & \textbf{0.00} & 0.02 & 34.39 & 6.09 & \textbf{0.00} & 0.08 & 59.23 \\
CASE\_22 & 10.02 & 11.05 & 12.31 & 9.80 & 1220.4 & 11.21 & \textbf{0.00} & 0.21 & 153.65 & 11.22 & 0.06 & 0.12 & 137.17 & 11.21 & 0.02 & 0.14 & 116.34 \\
CASE\_23 & 14.18 & 15.70 & 17.31 & 9.27 & 4356 & 15.86 & 0.11 & 0.21 & 222.80 & 15.87 & 0.15 & 0.18 & 197.95 & 15.84 & \textbf{0.00} & 0.12 & 163.00 \\
CASE\_24 & 19.27 & 21.26 & 23.02 & 7.51 & 10875.6 & 21.42 & 0.04 & 0.15 & 263.55 & 21.41 & \textbf{0.00} & 0.08 & 269.12 & 21.42 & 0.03 & 0.13 & 167.02 \\ \hline
\multicolumn{1}{r}{Averages} & \multicolumn{1}{l}{} & \multicolumn{1}{l}{} & \multicolumn{1}{l}{} & 5.29 & 4828.35 & \multicolumn{1}{l}{} & 0.03 & 0.08 & 155.76 & \multicolumn{1}{l}{} & 0.03 & 0.07 & 150.91 & \multicolumn{1}{l}{} & 0.05 & 0.12 & 87.37 \\ \hline
\end{tabular}%
}
\end{table}

\autoref{tab:resultLitHeuristicWithTW} shows our heuristic results with the literature instances with business hours. Similar to the previous table, the HSA results are reported from \cite{burdett2018integFJSS} with its solution times in seconds, makespan values for the best solution found (best) in days, its best and average relative percentile deviation, respectively BRPD and ARPD, and the average time to best (ATTB) in seconds. 
Upon analysis of the results, we observed that the RKO methods consistently enhanced the results for all instances from the literature. Specifically, SA found eight new best-known solutions (BKS), ILS found two new BKS, and BRKGA-QL found 18 new BKS.
Statistical tests confirm that all RKO heuristics demonstrated superior performance compared to the HSA, as shown in \autoref{tab:resultLitHeuristicNoTW}. An inverse performance order was observed among the RKO heuristics: BRKGA-QL outperformed the other two methods (Wilcoxon test $p$-values: 3.08E-08 when compared to SA, and 3.45E-10 when compared to ILS). No statistically significant difference was found between ILS and SA ($p$-value: 0.0528). Notably, the disparity between the ARPDs was more pronounced than in \autoref{tab:resultLitHeuristicNoTW}.


\begin{table}[htbp]
\renewcommand{\arraystretch}{1.1}
\centering
\caption{Literature instances heuristic results with business hours.}
\label{tab:resultLitHeuristicWithTW}
\resizebox{\textwidth}{!}{%
\begin{tabular}{l|rrr|rrrr|rrrr|rrrr} \hline
 & \multicolumn{3}{c|}{HSA} & \multicolumn{4}{c|}{SA} & \multicolumn{4}{c|}{ILS} & \multicolumn{4}{c}{BRKGA-QL} \\ \hline
\multicolumn{1}{c|}{Instance} & \multicolumn{1}{c}{best} & \multicolumn{1}{c}{BRPD(\%)} & \multicolumn{1}{c|}{time (s)} & \multicolumn{1}{c}{best} & \multicolumn{1}{c}{BRPD(\%)} & \multicolumn{1}{c}{ARPD(\%)} & \multicolumn{1}{c|}{ATTB(s)} & \multicolumn{1}{c}{best} & \multicolumn{1}{c}{BRPD(\%)} & \multicolumn{1}{c}{ARPD(\%)} & \multicolumn{1}{c|}{ATTB(s)} & \multicolumn{1}{c}{best} & \multicolumn{1}{c}{BRPD(\%)} & \multicolumn{1}{c}{ARPD(\%)} & \multicolumn{1}{c}{ATTB(s)} \\ \hline
CASE\_01 & 19.63 & 11.83 & 900 & 17.55 & \textbf{0.00} & 1.41 & 105.88 & 18.29 & 4.19 & 4.22 & 85.10 & 18.24 & 3.91 & 4.13 & 63.02 \\
CASE\_02 & 42.19 & 10.01 & 9432 & 39.35 & 2.61 & 2.77 & 182.27 & 38.51 & 0.42 & 1.91 & 166.37 & 38.35 & \textbf{0.00} & 1.58 & 128.39 \\
CASE\_03 & 59.33 & 3.45 & 32796 & 57.35 & \textbf{0.00} & 0.97 & 166.53 & 57.40 & 0.09 & 1.13 & 197.60 & 57.42 & 0.12 & 0.84 & 152.11 \\
CASE\_04 & 79.46 & 1.15 & 110160 & 79.39 & 1.06 & 2.13 & 261.99 & 79.33 & 0.98 & 1.69 & 239.34 & 78.55 & \textbf{0.00} & 1.15 & 170.95 \\
CASE\_05 & 20.01 & 13.42 & 828 & 17.64 & \textbf{0.00} & 0.26 & 83.20 & 17.65 & 0.02 & 2.48 & 68.14 & 17.69 & 0.29 & 2.83 & 62.50 \\
CASE\_06 & 38.59 & 9.02 & 8100 & 36.65 & 3.53 & 3.73 & 148.24 & 36.71 & 3.71 & 3.87 & 82.74 & 35.40 & \textbf{0.00} & 2.68 & 127.25 \\
CASE\_07 & 56.85 & 8.48 & 32652 & 54.16 & 3.35 & 4.55 & 165.40 & 53.35 & 1.80 & 3.24 & 168.99 & 52.40 & \textbf{0.00} & 1.83 & 164.81 \\
CASE\_08 & 68.41 & 1.25 & 13572 & 67.57 & \textbf{0.00} & 0.01 & 263.05 & 67.57 & 0.001 & 0.01 & 232.98 & 67.57 & 0.01 & 0.02 & 49.77 \\
CASE\_09 & 10.06 & 16.18 & 576 & 8.66 & \textbf{0.00} & 0.00 & 78.00 & 8.66 & \textbf{0.00} & 1.17 & 66.21 & 8.66 & \textbf{0.00} & 0.97 & 69.93 \\
CASE\_10 & 20.52 & 15.76 & 6120 & 18.43 & 3.98 & 4.59 & 163.51 & 18.38 & 3.69 & 4.22 & 145.38 & 17.73 & \textbf{0.00} & 3.53 & 88.51 \\
CASE\_11 & 32.08 & 21.38 & 24480 & 27.55 & 4.24 & 9.06 & 195.44 & 27.38 & 3.61 & 7.76 & 151.45 & 26.43 & \textbf{0.00} & 0.86 & 162.70 \\
CASE\_12 & 42.34 & 9.95 & 76464 & 39.50 & 2.57 & 4.41 & 246.59 & 38.81 & 0.79 & 2.00 & 271.43 & 38.51 & \textbf{0.00} & 1.53 & 281.31 \\
CASE\_13 & 11.65 & 28.89 & 504 & 9.04 & \textbf{0.00} & 1.69 & 86.49 & 9.34 & 3.32 & 3.96 & 51.51 & 9.34 & 3.32 & 3.33 & 71.88 \\
CASE\_14 & 21.41 & 16.68 & 5760 & 18.38 & 0.17 & 0.72 & 139.00 & 18.36 & 0.06 & 0.54 & 143.49 & 18.35 & \textbf{0.00} & 0.38 & 97.93 \\
CASE\_15 & 32.74 & 19.31 & 24012 & 29.39 & 7.10 & 7.30 & 174.32 & 29.34 & 6.92 & 7.22 & 165.69 & 27.44 & \textbf{0.00} & 1.33 & 208.53 \\
CASE\_16 & 44.43 & 10.39 & 69624 & 40.99 & 1.83 & 2.45 & 299.90 & 40.34 & 0.23 & 1.09 & 293.65 & 40.25 & \textbf{0.00} & 0.32 & 269.92 \\
CASE\_17 & 11.73 & 18.02 & 432 & 9.94 & \textbf{0.00} & 2.24 & 80.73 & 10.25 & 3.08 & 3.66 & 48.44 & 10.12 & 1.87 & 2.85 & 59.19 \\
CASE\_18 & 20.47 & 11.87 & 4716 & 18.44 & 0.79 & 0.95 & 175.08 & 18.41 & 0.59 & 0.81 & 189.10 & 18.30 & \textbf{0.00} & 0.37 & 106.51 \\
CASE\_19 & 33.41 & 18.00 & 20700 & 29.42 & 3.90 & 4.34 & 183.18 & 29.37 & 3.73 & 3.95 & 203.83 & 28.31 & \textbf{0.00} & 1.89 & 178.83 \\
CASE\_20 & 41.87 & 14.34 & 59904 & 37.53 & 2.47 & 4.53 & 267.41 & 37.53 & 2.48 & 4.34 & 269.61 & 36.62 & \textbf{0.00} & 2.48 & 223.29 \\
CASE\_21 & 7.88 & 22.08 & 360 & 6.45 & \textbf{0.00} & 0.16 & 69.78 & 6.45 & \textbf{0.00} & 0.31 & 36.21 & 6.45 & \textbf{0.00} & 0.06 & 51.74 \\
CASE\_22 & 15.15 & 22.36 & 4500 & 12.71 & 2.64 & 4.88 & 128.54 & 12.50 & 0.94 & 2.70 & 183.73 & 12.38 & \textbf{0.00} & 0.60 & 126.03 \\
CASE\_23 & 22.12 & 25.23 & 17928 & 18.33 & 3.80 & 4.79 & 218.44 & 18.32 & 3.72 & 4.06 & 204.68 & 17.66 & \textbf{0.00} & 2.37 & 190.68 \\
CASE\_24 & 27.36 & 12.32 & 56124 & 25.23 & 3.56 & 4.89 & 304.89 & 24.52 & 0.67 & 3.23 & 267.27 & 24.36 & \textbf{0.00} & 1.52 & 264.02 \\ \hline
Averages & \multicolumn{1}{l}{} & 14.22 & 24193.50 & \multicolumn{1}{l}{} & 1.98 & 3.03 & 174.49 & \multicolumn{1}{l}{} & 1.88 & 2.90 & 163.87 & \multicolumn{1}{l}{} & 0.40 & 1.64 & 140.41 \\ \hline
\end{tabular}%
}
\end{table}

\begin{figure}[htbp]
    \centering
    \includegraphics[width=0.65\linewidth]{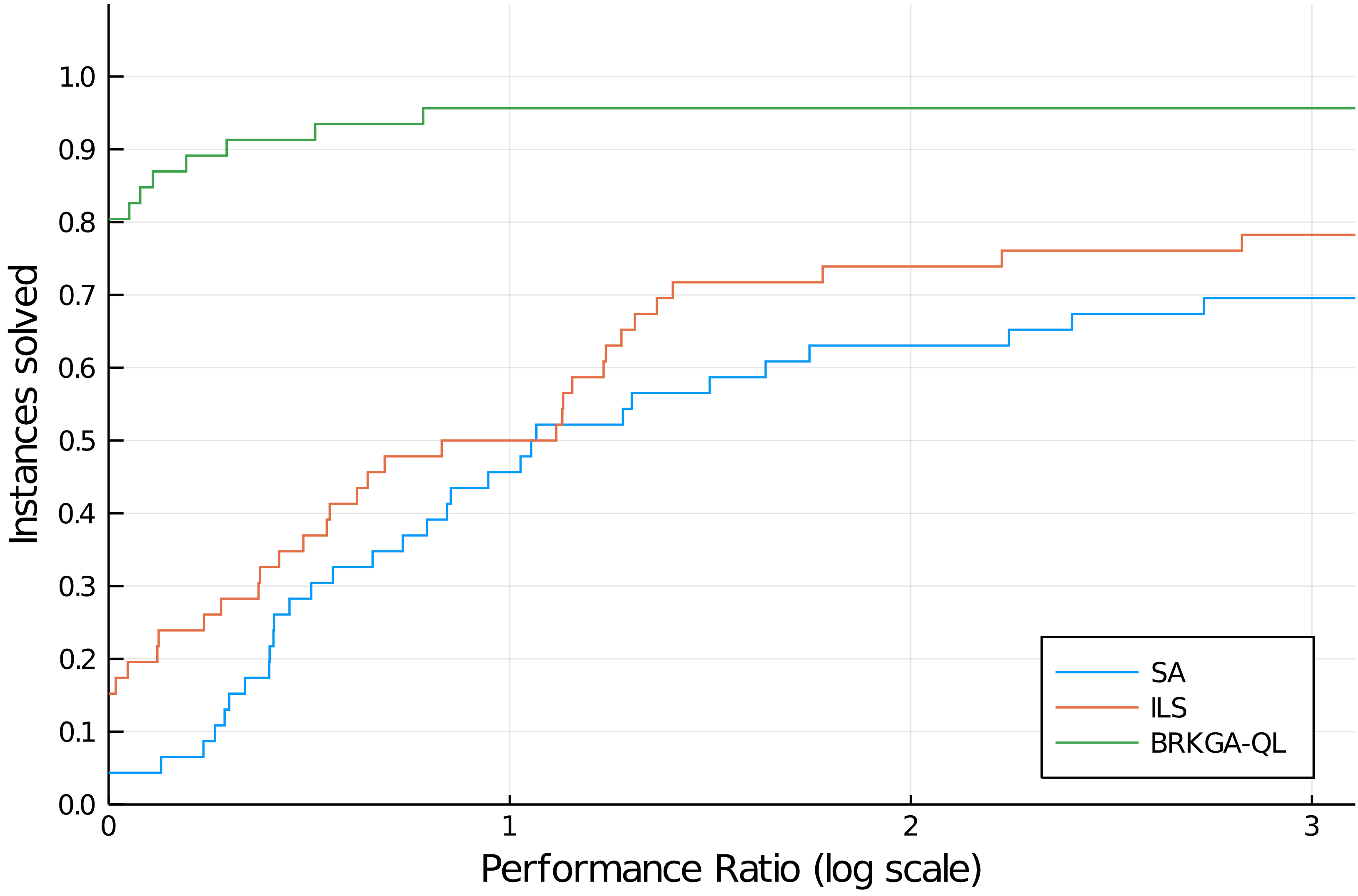}
    \caption{Performance Profile considering the RKO-based heuristics.}
    \label{fig:perfprof}
\end{figure}

\autoref{fig:perfprof} presents the performance profile proposed by \cite{DolanMore2002}, considering the RKO-based heuristics for BRKGA-QL, SA, and ILS, with a gap limit of $1\%$ relative to the best-known solutions (BKS) for each of the tested instances. When the performance ratio is zero, BRKGA-QL attains the best gaps for $80\%$ of the solved instances, ILS achieves the best gaps for $16\%$ of the instances, and SA achieves the best gaps for only $4\%$ of the solved instances. When the performance ratio is increased to 2 ($x$-axis $= log_2(2)=1$), BRKGA-QL solves $95\%$ of the instances with the best gaps, while ILS and SA solve $50\%$ and $46\%$ of the instances, respectively. As higher performance ratios are allowed, the percentage of instances solved with gaps smaller than $1\%$ remains constant for BRKGA-QL at $95\%$. In contrast, ILS and SA can solve $79\%$ and $70\% $ of the instances, respectively.

\begin{figure}[htbp]
    \centering
    \includegraphics[width=0.8\linewidth]{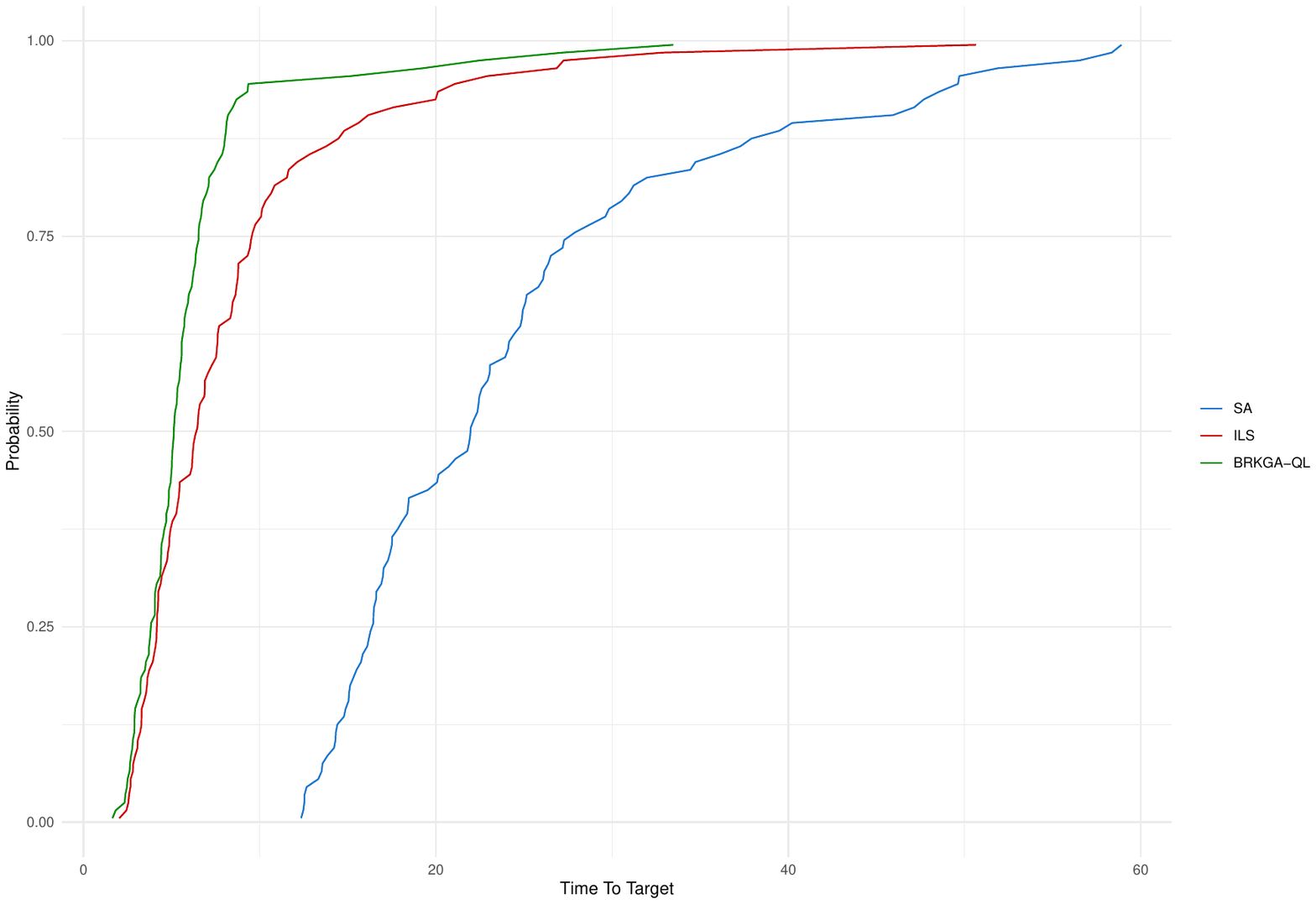}
    \caption{Time to target plot (Instance fjspnostr/CASE\_21)}
    \label{fig:ttt}
\end{figure}

The Time To Target (TTT) plot, as described in \cite{aiex2007ttt}, is used to analyze the RKO convergence rate considering BRKGA-QL, SA, and ILS heuristics. The instance set fjspnostr/CASE\_21 was considered for this test. All three solution procedures find the same solution, which is also the best-known solution for this instance. The experiment consists of running each method 100 times for the instance. Each run is independent and stops when a solution with a cost at least as good as a given target value is found. These experiments consider an integer value at most $0.50\%$ greater than the BKS solution as the target. The analysis of \autoref{fig:ttt} indicates that within the first $20$ seconds, BRKGA-QL has a $97\%$ probability of reaching the target value, compared to $93\%$ for ILS and $44\%$ for SA. By $40$ seconds, BRKGA-QL consistently achieves the target value, while ILS and SA reach probabilities of $98\%$ and $89\%$, respectively. The likelihood of reaching the target increases to $100\%$ at approximately $50$ seconds for ILS and $58$ seconds for SA. These results demonstrate that while all three methods exhibit strong convergence, BRKGA-QL achieves the fastest convergence.

Based on the computational tests, no single method outperformed the others in all cases. Therefore, we selected the BRKGA-QL method for the case study, as it provided better results for instances considering business hours. Additionally, this method does not require offline parameter tuning, making it more convenient for day-to-day hospital management. The case study instances are more complex and constrained by factors such as surgeon schedules, operating room types, equipment availability, and other additional attributes.


\subsection{Computational results for the case study instances}\label{sec:results-caseStudy}

Our case study instances were primarily generated using data collected from Santa Casa, a non-profit hospital with locations in several cities across Brazil. The surgery types, operating times, and probability distribution functions for occurrence were sourced from \cite{costa2017assessment} and \cite{siqueira2018caseBR}. The script initially receives or randomly picks some surgeries, multiplies per a constant, and picks a number inside a $\pm 15\%$ range for the number of surgeons, number of each room type, and available types of equipment. For a $14$-day planning horizon, operating rooms and surgeons are only available from 7:00 AM to 10:00 PM. The other room types are always available. Each day, a surgeon is available with a $70\%$ chance. Based on our case study collected data, each operating room is always available from Monday to Friday and from 7:00 AM to 1:00 PM on Saturdays with a $35\%$ chance. The equipment requirement is determined according to the surgery type. The generated set of instances is publicly available at a \href{https://github.com/brunosalezze/iorsp-instances}{\textit{Github} repository}.

\autoref{tab:resultCaseHeuristic} shows our heuristic results with the case study instances. The first tree columns describe the instances, followed by the best computed lower bound from \autoref{tab:resultCaseForm}, followed by the BRKGA-QL results split between the cases with and without availability slots (AS). For these instances, the tests without availability slots imply that all subjects (rooms, surgeons, patients, or equipment) are always available. Each instance has the best and average makespan value in days and the average CPU time in seconds. The case with no availability slots has a percentile optimality gap comparing the best solution to the best lower bound. 

Analysing the gap results, eight instances had gaps smaller than $1\%$, even among the larger ones. This shows that the BRKGA-QL found almost optimal results even in instances with many more constraints (surgeon schedules, OR compatibility, and equipment availability) in relation to the instances of \citep{burdett2018integFJSS}. BRKGA-QL demonstrated the ability to identify high-quality solutions within seconds of computation time, even for large-scale instances containing approximately 200 surgical procedures. The algorithm was executed 15 times for each instance, exhibiting robust performance by consistently producing average solutions close to the best solutions obtained. These attributes make BRKGA-QL suitable for practical implementation in hospital management systems, where computational efficiency and solution quality are essential.

\begin{table}[!h]
\centering
\caption{Case study instances heuristic results.}
\label{tab:resultCaseHeuristic}
\scalebox{0.67}{
\begin{tabular}{lrr|r|rrrrrrrrr}
\hline
         &           &           &         & \multicolumn{9}{c}{BRKGA-QL}                                                                                 \\ \cline{5-13} 
         &           &           &         & \multicolumn{5}{c|}{No AS}                                           & \multicolumn{4}{c}{With AS}           \\ \hline
Instance & Surgeries & Rooms     & Best LB & Best  & Average & ARPD(\%) & Time(s) & \multicolumn{1}{r|}{Gap (\%)} & Best   & Average & ARPD(\%) & Time(s) \\ \hline
p070     & 58        & 31,3,2,4  & 4.44    & 4.61  & 4.632   & 0.48     & 60.1    & \multicolumn{1}{r|}{3.85}     & 6.615  & 6.96    & 5.22     & 60.1    \\
p078     & 66        & 23,3,3,4  & 5.536   & 6.057 & 6.073   & 0.26     & 70.1    & \multicolumn{1}{r|}{9.41}     & 8.959  & 9.036   & 0.86     & 70.4    \\
p087     & 75        & 29,3,3,4  & 5.247   & 5.679 & 5.698   & 0.33     & 80.1    & \multicolumn{1}{r|}{8.23}     & 9.663  & 9.678   & 0.16     & 80.1    \\
p093     & 81        & 30,3,3,4  & 5.485   & 5.848 & 5.87    & 0.38     & 90.1    & \multicolumn{1}{r|}{6.62}     & 9.678  & 9.706   & 0.29     & 94.4    \\
p098     & 86        & 40,3,2,4  & 5.719   & 5.728 & 5.731   & 0.05     & 100.1   & \multicolumn{1}{r|}{0.15}     & 9.648  & 9.675   & 0.28     & 104.6   \\
p100     & 84        & 39,4,3,5  & 4.604   & 4.981 & 4.998   & 0.34     & 100.1   & \multicolumn{1}{r|}{8.19}     & 8.708  & 8.74    & 0.37     & 104.9   \\
p109     & 93        & 35,4,3,5  & 5.364   & 5.926 & 5.942   & 0.27     & 100.1   & \multicolumn{1}{r|}{10.47}    & 9.534  & 9.57    & 0.38     & 103.7   \\
p120     & 104       & 46,4,4,5  & 5.466   & 5.497 & 5.515   & 0.33     & 100.2   & \multicolumn{1}{r|}{0.57}     & 9.823  & 9.876   & 0.54     & 105     \\
p138     & 118       & 43,5,4,5  & 5.224   & 5.776 & 5.801   & 0.43     & 120.2   & \multicolumn{1}{r|}{10.55}    & 9.252  & 9.315   & 0.68     & 126     \\
p153     & 133       & 71,5,4,7  & 5.486   & 5.503 & 5.515   & 0.22     & 120.2   & \multicolumn{1}{r|}{0.3}      & 9.994  & 10.115  & 1.21     & 124.9   \\
p165     & 141       & 78,6,4,7  & 5.019   & 5.042 & 5.055   & 0.26     & 120.3   & \multicolumn{1}{r|}{0.46}     & 9.233  & 9.257   & 0.26     & 131.3   \\
p178     & 154       & 88,6,4,6  & 5.335   & 5.353 & 5.369   & 0.30     & 120.3   & \multicolumn{1}{r|}{0.34}     & 9.901  & 10.026  & 1.26     & 128.8   \\
p183     & 159       & 89,6,4,6  & 5.556   & 5.588 & 5.596   & 0.14     & 140.4   & \multicolumn{1}{r|}{0.56}     & 10.344 & 10.528  & 1.78     & 146.2   \\
p189     & 161       & 71,7,5,7  & 4.902   & 5.238 & 5.278   & 0.76     & 140.4   & \multicolumn{1}{r|}{6.86}     & 8.847  & 8.934   & 0.98     & 140.3   \\
p192     & 164       & 72,7,4,8  & 4.964   & 5.257 & 5.28    & 0.44     & 140.4   & \multicolumn{1}{r|}{5.9}      & 9.26   & 9.354   & 1.02     & 141.4   \\
p193     & 165       & 78,7,5,7  & 4.952   & 5.017 & 5.034   & 0.34     & 140.2   & \multicolumn{1}{r|}{1.3}      & 9.098  & 9.232   & 1.47     & 140.3   \\
p197     & 173       & 99,6,5,7  & 5.767   & 5.782 & 5.794   & 0.21     & 160.4   & \multicolumn{1}{r|}{0.25}     & 10.263 & 10.321  & 0.57     & 160.9   \\
p201     & 177       & 73,6,4,8  & 6.053   & 6.081 & 6.095   & 0.23     & 160.4   & \multicolumn{1}{r|}{0.46}     & 10.945 & 11.085  & 1.28     & 160.5   \\
p216     & 188       & 95,7,5,9  & 5.505   & 5.529 & 5.547   & 0.33     & 160.5   & \multicolumn{1}{r|}{0.44}     & 10.247 & 10.294  & 0.46     & 160.8   \\
p233     & 197       & 88,9,4,10 & 4.778   & 5.222 & 5.258   & 0.69     & 160.5   & \multicolumn{1}{r|}{9.27}     & 8.98   & 9.09    & 1.22     & 160     \\ \hline
\multicolumn{3}{l|}{Averages}    &         &       &         & 0.34     & 119.255 & \multicolumn{1}{r|}{4.20}     &        &         & 1.01     & 122.23  \\ \hline
\end{tabular}}
\end{table}


\subsection{Surgery fixing and rescheduling} \label{sec:surg_rescheduling}

Due to unforeseen reasons, a previously scheduled surgical procedure can be postponed or cancelled. This leads to surgery rescheduling, as highlighted by \cite{burdett2018integFJSS}. It is a challenging but necessary aspect of healthcare management. It requires balancing accommodating patient needs, ensuring patient safety, and efficiently managing resources. While some previously scheduled surgeries can be ultimately rescheduled, often, most of the surgeries closest to execution need to be maintained during the next scheduling. Our approach was developed with initial availability windows for every resource to allow a more flexible input. Some of the reserved initial schedules can also be pre-scheduled surgeries.

\begin{figure}[!ht]
    \centering
    \scalebox{0.45}{\includegraphics{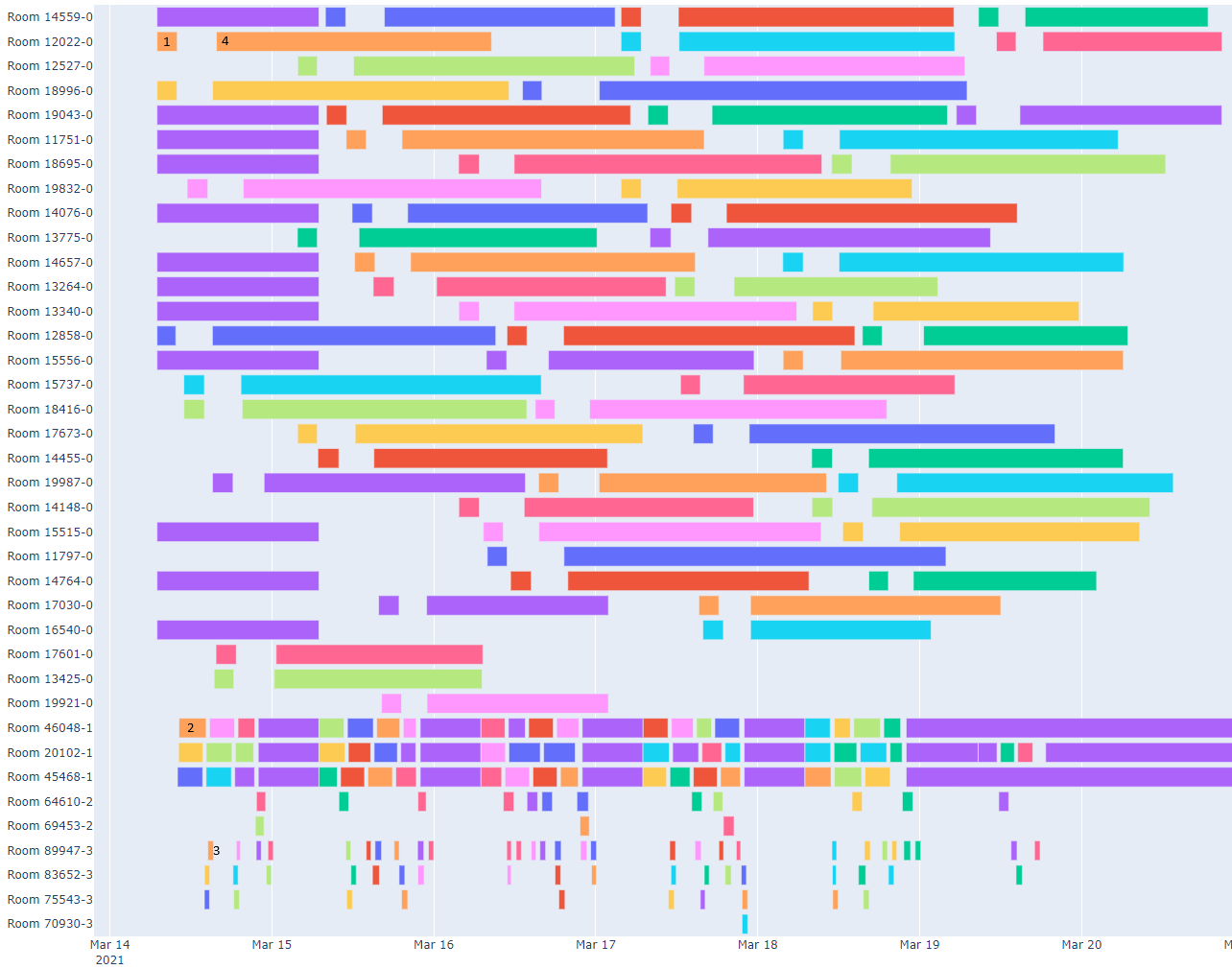}}
    \caption{Rooms schedule of the best-known solution to $p70$.}
    \label{fig:sol_p70_initial}
\end{figure}

As an example of how input can be used for rescheduling efforts, we selected the best-known solution to instance $p70$, shown in \autoref{fig:sol_p70_initial}, the non-available moments are marked with the colour purple $(171,99,250)$, on a 0-255 (red, green, blue) scale. We highlight the schedule of a patient as an example. The patient of colour orange $(255,161,90)$ has its events numbered. The patient arrives at the hospital and first goes to Room $12022-0$ (1), goes to surgery in Room $46048-1$ (2), leaves to observation in Room $89947-3$ (3) and moves to recovery back in Room $12022-0$ (4).

\begin{figure}[!ht]
    \centering
    \scalebox{0.5}{\includegraphics{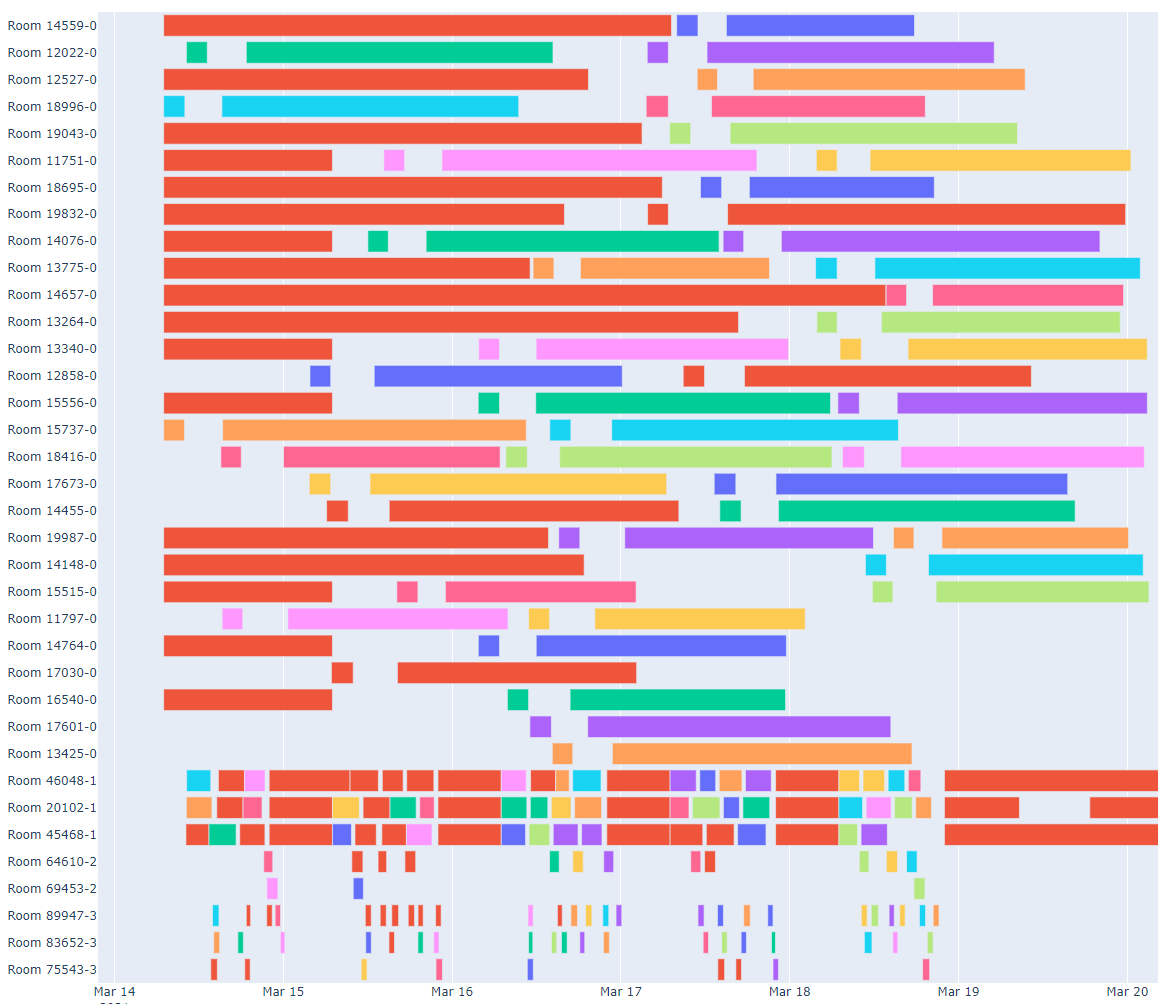}}
    \caption{Rooms schedule of a rescheduled solution to $p70$.}
    \label{fig:sol_p70_rescheduled}
\end{figure}

Out of the initial $58$ surgeries to be scheduled, we randomly removed $11$ to simulate cancelled surgeries, followed by the fixation of programmed schedules of $10$ other ones, fixing the rooms, surgeons, and types of equipment reserved for these time intervals. We selected mainly from the first to be performed for the initial scheduling. Finally, we created $7$ new surgeries. The BRKGA-QL was run with the new instance under the same conditions as when $p70$ was initially solved with ten runs. The room scheduling for the best solution found can be observed in Figure \autoref{fig:sol_p70_rescheduled}. The fixed time intervals are the same colour as the non-business hour intervals for all rooms, marked with red $(239,85,59)$. The algorithm could fulfil the ORs' time slots as efficiently as possible without pre-scheduled surgeries.

\section{Conclusions and future works} \label{sec:conclusions}

This research highlights the importance of effective surgery scheduling in managing healthcare institutions. The Random Key Optimizer (RKO), incorporating the BRKGA-QL, Simulated Annealing, and Iterated Local Search methods, marks a significant advancement in tackling the complex challenges of surgery scheduling. By carefully managing the availability of rooms, patients, and surgeons, our methods generate efficient schedules and support flexible re-scheduling, aligning with the dynamic demands of healthcare environments.

Using real-world data to generate and test new instances enhances the applicability and relevance of our approach. The introduction of simple lower-bound formulations serves as a critical contribution, enabling a nuanced evaluation of optimality gaps and providing a benchmark for assessing the effectiveness of the heuristic results. Considering the three metaheuristics, the statistical analysis highlights the RKO as a valuable tool for optimizing surgery scheduling in this case study. In this context, BRKGA-QL consistently demonstrates its potential to identify superior solutions, mainly when business hours are considered.

Our research showcases tangible improvements over existing literature instances, demonstrating superior performance across lower and upper bounds. Additionally, the efficiency in generating schedules for the newly introduced instances signifies our method's practical applicability and scalability.

This study contributes substantively to the ongoing discourse on optimizing surgery scheduling, offering practical solutions to healthcare practitioners and administrators. The insights gained from our research can empower hospitals to refine resource allocation, minimize patient wait times, and elevate overall operational efficiency. The proposed RKO approach and its associated formulations stand poised to make enduring contributions to the evolving landscape of healthcare scheduling and optimization.

For future works, we plan to investigate a complete mathematical model for the IORSP and a hybrid exact algorithm that could improve the solution quality and the bounds.

\section*{Acknowledgments}

This work was supported by the S\~ao Paulo Research Foundation (FAPESP) under grant \#2021/09482-4. Antonio A. Chaves was supported by FAPESP under grants \#2018/15417-8 and \#2022/05803-3, and National Council for Scientific and Technological Development (CNPq) under grants \#423694/2018-9 and \#303736/2018-6. Eduardo M. Silva was supported by the S\~ao Paulo Research Foundation (FAPESP) under grant \#2023/04588-4.

\bibliographystyle{plainnat}
\bibliography{referencias}

\begin{thebibliography}{58}
\providecommand{\natexlab}[1]{#1}
\providecommand{\url}[1]{\texttt{#1}}
\expandafter\ifx\csname urlstyle\endcsname\relax
  \providecommand{\doi}[1]{doi: #1}\else
  \providecommand{\doi}{doi: \begingroup \urlstyle{rm}\Url}\fi

\bibitem[Abdelrasol et~al.(2014)Abdelrasol, Harraz, and
  Eltawil]{abdelrasol2014orspSurv}
Zakaria Abdelrasol, Nermine Harraz, and Amr Eltawil.
\newblock Operating room scheduling problems: A survey and a proposed solution
  framework.
\newblock In Haeng~Kon Kim, Sio-Iong Ao, and Mahyar~A. Amouzegar, editors,
  \emph{Transactions on Engineering Technologies}, pages 717--731, Dordrecht,
  2014. Springer Netherlands.
\newblock ISBN 978-94-017-9115-1.

\bibitem[Aiex et~al.(2007)Aiex, Resende, and Ribeiro]{aiex2007ttt}
Renata~M Aiex, Mauricio~GC Resende, and Celso~C Ribeiro.
\newblock Ttt plots: a perl program to create time-to-target plots.
\newblock \emph{Optimization Letters}, 1\penalty0 (4):\penalty0 355--366, 2007.

\bibitem[Akbarzadeh et~al.(2020)Akbarzadeh, Moslehi, Reisi-Nafchi, and
  Maenhout]{akbarzadeh2020divHeur}
Babak Akbarzadeh, Ghasem Moslehi, Mohammad Reisi-Nafchi, and Broos Maenhout.
\newblock A diving heuristic for planning and scheduling surgical cases in the
  operating room department with nurse re-rostering.
\newblock \emph{Journal of Scheduling}, 23\penalty0 (2):\penalty0 265--288, Apr
  2020.
\newblock ISSN 1099-1425.
\newblock \doi{10.1007/s10951-020-00639-6}.
\newblock URL \url{https://doi.org/10.1007/s10951-020-00639-6}.

\bibitem[Andrade et~al.(2019)Andrade, Silva, and Pessoa]{ANDRADE201967}
Carlos~E. Andrade, Thuener Silva, and Luciana~S. Pessoa.
\newblock Minimizing flowtime in a flowshop scheduling problem with a biased
  random-key genetic algorithm.
\newblock \emph{Expert Systems with Applications}, 128:\penalty0 67--80, 2019.
\newblock ISSN 0957-4174.
\newblock \doi{https://doi.org/10.1016/j.eswa.2019.03.007}.
\newblock URL
  \url{https://www.sciencedirect.com/science/article/pii/S0957417419301605}.

\bibitem[Aringhieri et~al.(2015)Aringhieri, Landa, Soriano, Tànfani, and
  Testi]{aringhieri2015twolevel}
Roberto Aringhieri, Paolo Landa, Patrick Soriano, Elena Tànfani, and Angela
  Testi.
\newblock A two level metaheuristic for the operating room scheduling and
  assignment problem.
\newblock \emph{Computers \& Operations Research}, 54:\penalty0 21--34, 2015.
\newblock ISSN 0305-0548.
\newblock \doi{https://doi.org/10.1016/j.cor.2014.08.014}.
\newblock URL
  \url{https://www.sciencedirect.com/science/article/pii/S030505481400224X}.

\bibitem[Baxter(1981)]{baxter1981local}
J.~Baxter.
\newblock Local optima avoidance in depot location.
\newblock \emph{Journal of the Operational Research Society}, 32:\penalty0
  815--819, 1981.

\bibitem[Bean(1994)]{bean1994rkga}
James~C. Bean.
\newblock Genetic algorithms and random keys for sequencing and optimization.
\newblock \emph{ORSA Journal on Computing}, 6\penalty0 (2):\penalty0 154--160,
  1994.
\newblock \doi{10.1287/ijoc.6.2.154}.
\newblock URL \url{https://doi.org/10.1287/ijoc.6.2.154}.

\bibitem[Brucker and Schlie(1990)]{brucker1990firstfjss}
P.~Brucker and R.~Schlie.
\newblock Job-shop scheduling with multi-purpose machines.
\newblock \emph{Computing}, 45\penalty0 (4):\penalty0 369--375, Dec 1990.
\newblock ISSN 1436-5057.
\newblock \doi{10.1007/BF02238804}.
\newblock URL \url{https://doi.org/10.1007/BF02238804}.

\bibitem[Burdett and Kozan(2018)]{burdett2018integFJSS}
Robert~L. Burdett and Erhan Kozan.
\newblock An integrated approach for scheduling health care activities in a
  hospital.
\newblock \emph{European Journal of Operational Research}, 264\penalty0
  (2):\penalty0 756--773, 2018.
\newblock ISSN 0377-2217.
\newblock \doi{https://doi.org/10.1016/j.ejor.2017.06.051}.
\newblock URL
  \url{https://www.sciencedirect.com/science/article/pii/S0377221717305921}.

\bibitem[Chaudhry and Khan(2016)]{chaudhry2016fjssreview}
Imran~Ali Chaudhry and Abid~Ali Khan.
\newblock A research survey: review of flexible job shop scheduling techniques.
\newblock \emph{International Transactions in Operational Research},
  23\penalty0 (3):\penalty0 551--591, 2016.
\newblock \doi{https://doi.org/10.1111/itor.12199}.
\newblock URL \url{https://onlinelibrary.wiley.com/doi/abs/10.1111/itor.12199}.

\bibitem[Chaves et~al.(2024)Chaves, Resende, and
  Silva]{chaves2024randomkeygraspcombinatorialoptimization}
Antonio~A. Chaves, Mauricio G.~C. Resende, and Ricardo M.~A. Silva.
\newblock A random-key grasp for combinatorial optimization, 2024.
\newblock URL \url{https://arxiv.org/abs/2405.18681}.

\bibitem[Chaves and Lorena(2021)]{chaves2021brkgaql}
Antônio~Augusto Chaves and Luiz Henrique~Nogueira Lorena.
\newblock An adaptive and near parameter-free brkga using q-learning method.
\newblock In \emph{2021 IEEE Congress on Evolutionary Computation (CEC)}, pages
  2331--2338, 2021.
\newblock \doi{10.1109/CEC45853.2021.9504766}.

\bibitem[Costa(2017)]{costa2017assessment}
Altair da~Silva Costa.
\newblock Assessment of operative times of multiple surgical specialties in a
  public university hospital.
\newblock \emph{Einstein (Sao Paulo)}, 15:\penalty0 200--205, 2017.
\newblock \doi{10.1590/S1679-45082017GS3902}.

\bibitem[Dang et~al.(2017)Dang, C\'{a}ceres, De~Causmaecker, and
  St\"{u}tzle]{dang2017irace}
Nguyen Dang, Leslie~P\'{e}rez C\'{a}ceres, Patrick De~Causmaecker, and Thomas
  St\"{u}tzle.
\newblock Configuring irace using surrogate configuration benchmarks.
\newblock In \emph{GECCO '17: Proceedings of the Genetic and Evolutionary
  Computation Conference}, GECCO '17, page 243–250, New York, NY, USA, 2017.
  Association for Computing Machinery.
\newblock ISBN 9781450349208.
\newblock \doi{10.1145/3071178.3071238}.
\newblock URL \url{https://doi.org/10.1145/3071178.3071238}.

\bibitem[Davis(1991)]{davis1991handbook}
Lawrence~D. Davis, editor.
\newblock \emph{Handbook of Genetic Algorithms}.
\newblock Van Nostrand Reinhold, New York, 1991.

\bibitem[Dekkers and Aarts(1991)]{Dekkers1991}
A.~Dekkers and E.~Aarts.
\newblock Global optimization and simulated annealing.
\newblock \emph{Mathematical Programming}, 50\penalty0 (3):\penalty0 367--393,
  1991.

\bibitem[Dellaert and Jeunet(2017)]{dellaert2017vnsSTP}
Nico Dellaert and Jully Jeunet.
\newblock A variable neighborhood search algorithm for the surgery tactical
  planning problem.
\newblock \emph{Computers \& Operations Research}, 84:\penalty0 216--225, 2017.
\newblock ISSN 0305-0548.
\newblock \doi{https://doi.org/10.1016/j.cor.2016.05.013}.
\newblock URL
  \url{https://www.sciencedirect.com/science/article/pii/S0305054816301204}.

\bibitem[Dexter et~al.(1999)Dexter, Macario, and Traub]{dexter1999binpack}
Franklin Dexter, Alex Macario, and Rodney D. Traub.
\newblock {Which Algorithm for Scheduling Add-on Elective Cases Maximizes
  Operating Room Utilization: Use of Bin Packing Algorithms and Fuzzy
  Constraints in Operating Room Management}.
\newblock \emph{Anesthesiology}, 91\penalty0 (5):\penalty0 1491--1491, 11 1999.
\newblock ISSN 0003-3022.
\newblock \doi{10.1097/00000542-199911000-00043}.
\newblock URL \url{https://doi.org/10.1097/00000542-199911000-00043}.

\bibitem[{Dolan} and {Mor\'{e}}(2002)]{DolanMore2002}
E.~D. {Dolan} and J.~J. {Mor\'{e}}.
\newblock Benchmarking optimization software with performance profiles.
\newblock \emph{Mathematical Programming}, 91\penalty0 (2):\penalty0 201 --
  213, 2002.

\bibitem[Dur{\'a}n et~al.(2017)Dur{\'a}n, Rey, and Wolff]{duran2017prioriLists}
Guillermo Dur{\'a}n, Pablo~A. Rey, and Patricio Wolff.
\newblock Solving the operating room scheduling problem with prioritized lists
  of patients.
\newblock \emph{Annals of Operations Research}, 258\penalty0 (2):\penalty0
  395--414, Nov 2017.
\newblock ISSN 1572-9338.
\newblock \doi{10.1007/s10479-016-2172-x}.
\newblock URL \url{https://doi.org/10.1007/s10479-016-2172-x}.

\bibitem[Fei et~al.(2009)Fei, Chu, and Meskens]{fei2009tatical}
H.~Fei, C.~Chu, and N.~Meskens.
\newblock Solving a tactical operating room planning problem
  by a column-generation-based heuristic procedure with four criteria.
\newblock \emph{Annals of Operations Research}, 166\penalty0 (1):\penalty0
  91--108, Feb 2009.
\newblock ISSN 1572-9338.
\newblock \doi{10.1007/s10479-008-0413-3}.
\newblock URL \url{https://doi.org/10.1007/s10479-008-0413-3}.

\bibitem[Fei et~al.(2010)Fei, Meskens, and Chu]{fei2010planning}
H.~Fei, N.~Meskens, and C.~Chu.
\newblock A planning and scheduling problem for an operating theatre using an
  open scheduling strategy.
\newblock \emph{Computers \& Industrial Engineering}, 58\penalty0 (2):\penalty0
  221--230, 2010.
\newblock ISSN 0360-8352.
\newblock \doi{https://doi.org/10.1016/j.cie.2009.02.012}.
\newblock URL
  \url{https://www.sciencedirect.com/science/article/pii/S0360835209000801}.
\newblock Scheduling in Healthcare and Industrial Systems.

\bibitem[Gonçalves and Resende(2011)]{goncalves2011brkga}
J.~F. Gonçalves and M.~G.~C. Resende.
\newblock Biased random-key genetic algorithms for combinatorial optimization.
\newblock \emph{Journal of Heuristics}, 17\penalty0 (5):\penalty0 487--525,
  2011.

\bibitem[{Gurobi Optimization, LLC}(2023)]{gurobi}
{Gurobi Optimization, LLC}.
\newblock {Gurobi Optimizer Reference Manual}, 2023.
\newblock URL \url{https://www.gurobi.com}.

\bibitem[Hamid et~al.(2019)Hamid, Nasiri, Werner, Sheikhahmadi, and
  Zhalechian]{hamid2019teamMemb}
Mahdi Hamid, Mohammad~Mahdi Nasiri, Frank Werner, Farrokh Sheikhahmadi, and
  Mohammad Zhalechian.
\newblock Operating room scheduling by considering the decision-making styles
  of surgical team members: A comprehensive approach.
\newblock \emph{Computers \& Operations Research}, 108:\penalty0 166--181,
  2019.
\newblock ISSN 0305-0548.
\newblock \doi{https://doi.org/10.1016/j.cor.2019.04.010}.
\newblock URL
  \url{https://www.sciencedirect.com/science/article/pii/S0305054819300930}.

\bibitem[Karafotias et~al.(2015)Karafotias, Hoogendoorn, and
  Eiben]{karafotias2015evaluating}
Giorgos Karafotias, Mark Hoogendoorn, and A.~E. Eiben.
\newblock Evaluating reward definitions for parameter control.
\newblock In Antonio~M. Mora and Giovanni Squillero, editors,
  \emph{Applications of Evolutionary Computation}, pages 667--680, Cham, 2015.
  Springer International Publishing.
\newblock ISBN 978-3-319-16549-3.

\bibitem[Kirkpatrick et~al.(1983)Kirkpatrick, Gelatt, and
  Vecchi]{Kirkpatrick1983}
S.~Kirkpatrick, C.~D. Gelatt, and M.~P. Vecchi.
\newblock Optimization by simulated annealing.
\newblock \emph{Science}, 220\penalty0 (4598):\penalty0 671--680, 1983.

\bibitem[Landa et~al.(2016)Landa, Aringhieri, Soriano, Tànfani, and
  Testi]{landa2016hybrid}
Paolo Landa, Roberto Aringhieri, Patrick Soriano, Elena Tànfani, and Angela
  Testi.
\newblock A hybrid optimization algorithm for surgeries scheduling.
\newblock \emph{Operations Research for Health Care}, 8:\penalty0 103--114,
  2016.
\newblock ISSN 2211-6923.
\newblock \doi{https://doi.org/10.1016/j.orhc.2016.01.001}.
\newblock URL
  \url{https://www.sciencedirect.com/science/article/pii/S2211692315300084}.

\bibitem[Lin and Li(2021)]{lin2021orspArtBee}
Yang-Kuei Lin and Min-Yang Li.
\newblock Solving operating room scheduling problem using artificial bee colony
  algorithm.
\newblock \emph{Healthcare}, 9\penalty0 (2), 2021.
\newblock ISSN 2227-9032.
\newblock \doi{10.3390/healthcare9020152}.
\newblock URL \url{https://www.mdpi.com/2227-9032/9/2/152}.

\bibitem[Liu et~al.(2011)Liu, Chu, and Wang]{liu2011heurORSP}
Ya~Liu, Chengbin Chu, and Kanliang Wang.
\newblock A new heuristic algorithm for the operating room scheduling problem.
\newblock \emph{Computers \& Industrial Engineering}, 61\penalty0 (3):\penalty0
  865--871, 2011.
\newblock ISSN 0360-8352.
\newblock \doi{https://doi.org/10.1016/j.cie.2011.05.020}.
\newblock URL
  \url{https://www.sciencedirect.com/science/article/pii/S0360835211001458}.

\bibitem[Lourenço et~al.(2003)Lourenço, Martin, and
  Stützle]{lourenco2003iterated}
H.~R. Lourenço, O.~Martin, and T.~Stützle.
\newblock Iterated local search.
\newblock In \emph{Handbook of Metaheuristics}, International Series in
  Operations Research \& Management Science, pages 321--353. Kluwer Academic
  Publishers, 2003.
\newblock ISBN 978-1-4020-7263-5.

\bibitem[Mangussi et~al.(2023)Mangussi, Pola, Macedo, ao, Proen\c{c}a,
  Gianfelice, Salezze, and Chaves]{mangussi2023meta}
A.~D. Mangussi, H.~Pola, H.~G. Macedo, L.~A.~Juli\ ao, M.~P.~T. Proen\c{c}a,
  P.~R.~L. Gianfelice, B.~V. Salezze, and A.~A. Chaves.
\newblock Meta-heur{\'i}sticas via chaves aleat{\'o}rias aplicadas ao problema
  de localiza{\c c}{\~a}o de hubs em {\'a}rvore.
\newblock In \emph{Anais do Simp{\'o}sio Brasileiro de Pesquisa Operacional},
  S{\~a}o Jos{\'e} dos Campos, 2023. Galo{\'a}.

\bibitem[Marques and Captivo(2015)]{marques2015orElective}
Inês Marques and M.~Eugénia Captivo.
\newblock Bicriteria elective surgery scheduling using an evolutionary
  algorithm.
\newblock \emph{Operations Research for Health Care}, 7:\penalty0 14--26, 2015.
\newblock ISSN 2211-6923.
\newblock \doi{https://doi.org/10.1016/j.orhc.2015.07.004}.
\newblock URL
  \url{https://www.sciencedirect.com/science/article/pii/S2211692314200580}.
\newblock ORAHS 2014 - The 40th international conference of the EURO working
  group on Operational Research Applied to Health Services.

\bibitem[Metropolis et~al.(1953)Metropolis, Rosenbluth, Rosenbluth, Teller, and
  Teller]{metropolis1953}
Nicholas Metropolis, Arianna Rosenbluth, Marshall Rosenbluth, Augusta Teller,
  and Edward Teller.
\newblock Equation of state calculations by fast computing machines.
\newblock \emph{Journal of Chemical Physics}, 21:\penalty0 1087--1092, 1953.

\bibitem[Mladenović and Hansen(1997)]{MLADENOVIC19971097}
N.~Mladenović and P.~Hansen.
\newblock Variable neighborhood search.
\newblock \emph{Computers \& Operations Research}, 24\penalty0 (11):\penalty0
  1097--1100, 1997.
\newblock ISSN 0305-0548.
\newblock \doi{https://doi.org/10.1016/S0305-0548(97)00031-2}.
\newblock URL
  \url{https://www.sciencedirect.com/science/article/pii/S0305054897000312}.

\bibitem[Molina-Pariente et~al.(2015)Molina-Pariente, Fernandez-Viagas, and
  Framinan]{molina2015integrated}
Jose~M. Molina-Pariente, Victor Fernandez-Viagas, and Jose~M. Framinan.
\newblock Integrated operating room planning and scheduling problem with
  assistant surgeon dependent surgery durations.
\newblock \emph{Computers \& Industrial Engineering}, 82:\penalty0 8--20, 2015.
\newblock ISSN 0360-8352.
\newblock \doi{https://doi.org/10.1016/j.cie.2015.01.006}.
\newblock URL
  \url{https://www.sciencedirect.com/science/article/pii/S0360835215000157}.

\bibitem[Nelder and Mead(1965)]{nelder1965simplex}
J.~A. Nelder and R.~Mead.
\newblock A simplex method for function minimization.
\newblock \emph{The Computer Journal}, 7\penalty0 (4):\penalty0 308--313, 1965.

\bibitem[Niven et~al.(1991)Niven, Zuckerman, and
  Montgomery]{niven1991introduction}
I.~Niven, H.~S. Zuckerman, and H.~L. Montgomery.
\newblock \emph{An Introduction to the Theory of Numbers}.
\newblock John Wiley \& Sons, 1991.

\bibitem[Ozkarahan(1995)]{ozkarahan1995alloc}
Irem Ozkarahan.
\newblock Allocation of surgical procedures to operating rooms.
\newblock \emph{Journal of Medical Systems}, 19\penalty0 (4):\penalty0
  333--352, Aug 1995.
\newblock ISSN 1573-689X.
\newblock \doi{10.1007/BF02257264}.
\newblock URL \url{https://doi.org/10.1007/BF02257264}.

\bibitem[Park et~al.(2021)Park, Kim, Eom, and Choi]{park2021orspPref}
Jaesang Park, Byung-In Kim, Myungeun Eom, and Byung~Kwan Choi.
\newblock Operating room scheduling considering surgeons’ preferences and
  cooperative operations.
\newblock \emph{Computers \& Industrial Engineering}, 157:\penalty0 107306,
  2021.
\newblock ISSN 0360-8352.
\newblock \doi{https://doi.org/10.1016/j.cie.2021.107306}.
\newblock URL
  \url{https://www.sciencedirect.com/science/article/pii/S0360835221002102}.

\bibitem[Rahimi and Gandomi(2021)]{rahimi2021review}
Iman Rahimi and Amir~H. Gandomi.
\newblock A comprehensive review and analysis of operating room and surgery
  scheduling.
\newblock \emph{Archives of Computational Methods in Engineering}, 28\penalty0
  (3):\penalty0 1667--1688, May 2021.
\newblock ISSN 1886-1784.
\newblock \doi{10.1007/s11831-020-09432-2}.
\newblock URL \url{https://doi.org/10.1007/s11831-020-09432-2}.

\bibitem[Roshanaei et~al.(2017)Roshanaei, Luong, Aleman, and
  Urbach]{roshanaei2017colaborative}
Vahid Roshanaei, Curtiss Luong, Dionne~M. Aleman, and David~R. Urbach.
\newblock Collaborative operating room planning and scheduling.
\newblock \emph{INFORMS Journal on Computing}, 29\penalty0 (3):\penalty0
  558--580, 2017.
\newblock \doi{10.1287/ijoc.2017.0745}.
\newblock URL \url{https://doi.org/10.1287/ijoc.2017.0745}.

\bibitem[Roshanaei et~al.(2020)Roshanaei, Booth, Aleman, Urbach, and
  Beck]{roshanaei2020branchCheck}
Vahid Roshanaei, Kyle~E.C. Booth, Dionne~M. Aleman, David~R. Urbach, and
  J.~Christopher Beck.
\newblock Branch-and-check methods for multi-level operating room planning and
  scheduling.
\newblock \emph{International Journal of Production Economics}, 220:\penalty0
  107433, 2020.
\newblock ISSN 0925-5273.
\newblock \doi{https://doi.org/10.1016/j.ijpe.2019.07.006}.
\newblock URL
  \url{https://www.sciencedirect.com/science/article/pii/S0925527319302439}.

\bibitem[Schuetz et~al.(2022)Schuetz, Brubaker, Montagu, van Dijk, Klepsch,
  Ross, Luckow, Resende, and Katzgraber]{schuetz2022rko}
Martin~J.A. Schuetz, J.~Kyle Brubaker, Henry Montagu, Yannick van Dijk,
  Johannes Klepsch, Philipp Ross, Andre Luckow, Mauricio~G.C. Resende, and
  Helmut~G. Katzgraber.
\newblock Optimization of robot-trajectory planning with nature-inspired and
  hybrid quantum algorithms.
\newblock \emph{Phys. Rev. Appl.}, 18:\penalty0 054045, Nov 2022.
\newblock \doi{10.1103/PhysRevApplied.18.054045}.
\newblock URL \url{https://link.aps.org/doi/10.1103/PhysRevApplied.18.054045}.

\bibitem[Siqueira et~al.(2018{\natexlab{a}})Siqueira, Arruda, Bahiense, Bahr,
  and Motta]{siqueira2018caseBR}
Cecília~L. Siqueira, Edilson~F. Arruda, Laura Bahiense, Germana~L. Bahr, and
  Geraldo~R. Motta.
\newblock Long-term integrated surgery room optimization and recovery ward
  planning, with a case study in the brazilian national institute of
  traumatology and orthopedics (into).
\newblock \emph{European Journal of Operational Research}, 264\penalty0
  (3):\penalty0 870--883, 2018{\natexlab{a}}.
\newblock ISSN 0377-2217.
\newblock \doi{https://doi.org/10.1016/j.ejor.2016.09.021}.
\newblock URL
  \url{https://www.sciencedirect.com/science/article/pii/S0377221716307573}.

\bibitem[Siqueira et~al.(2018{\natexlab{b}})Siqueira, Arruda, Bahiense, Bahr,
  and Motta]{siqueira2018longInt}
Cecília~L. Siqueira, Edilson~F. Arruda, Laura Bahiense, Germana~L. Bahr, and
  Geraldo~R. Motta.
\newblock Long-term integrated surgery room optimization and recovery ward
  planning, with a case study in the brazilian national institute of
  traumatology and orthopedics (into).
\newblock \emph{European Journal of Operational Research}, 264\penalty0
  (3):\penalty0 870--883, 2018{\natexlab{b}}.
\newblock ISSN 0377-2217.
\newblock \doi{https://doi.org/10.1016/j.ejor.2016.09.021}.
\newblock URL
  \url{https://www.sciencedirect.com/science/article/pii/S0377221716307573}.

\bibitem[Spears and De~Jong(1991)]{spears1991pux}
William Spears and Kenneth De~Jong.
\newblock On the virtues of parametrized uniform crossover.
\newblock In \emph{ICGA}, pages 230--236, 01 1991.

\bibitem[Subramanian et~al.(2010)Subramanian, Drummond, Bentes, Ochi, and
  Farias]{subramanian2010parallel}
Anand Subramanian, L{\'u}cia~MA Drummond, Cristiana Bentes, Luiz~Satoru Ochi,
  and Ricardo Farias.
\newblock A parallel heuristic for the vehicle routing problem with
  simultaneous pickup and delivery.
\newblock \emph{Computers \& Operations Research}, 37\penalty0 (11):\penalty0
  1899--1911, 2010.

\bibitem[Sutton and Barto(1999)]{sutton1999reinforcement}
Richard~S. Sutton and Andrew Barto.
\newblock {Reinforcement Learning}.
\newblock \emph{Journal of Cognitive Neuroscience}, 11\penalty0 (1):\penalty0
  126--134, 01 1999.
\newblock ISSN 0898-929X.
\newblock \doi{10.1162/089892999563184}.
\newblock URL \url{https://doi.org/10.1162/089892999563184}.

\bibitem[{Thomas Schneider} et~al.(2020){Thomas Schneider}, {Theresia van
  Essen}, Carlier, and Hans]{schneider2020downRes}
A.J. {Thomas Schneider}, J.~{Theresia van Essen}, Mijke Carlier, and Erwin~W.
  Hans.
\newblock Scheduling surgery groups considering multiple downstream resources.
\newblock \emph{European Journal of Operational Research}, 282\penalty0
  (2):\penalty0 741--752, 2020.
\newblock ISSN 0377-2217.
\newblock \doi{https://doi.org/10.1016/j.ejor.2019.09.029}.
\newblock URL
  \url{https://www.sciencedirect.com/science/article/pii/S0377221719307854}.

\bibitem[Vali et~al.(2022)Vali, Salimifard, Gandomi, and
  Chaussalet]{vali2022aplicFJSS}
Masoumeh Vali, Khodakaram Salimifard, Amir~H. Gandomi, and Thierry~J.
  Chaussalet.
\newblock Application of job shop scheduling approach in green patient flow
  optimization using a hybrid swarm intelligence.
\newblock \emph{Computers \& Industrial Engineering}, 172:\penalty0 108603,
  2022.
\newblock ISSN 0360-8352.
\newblock \doi{https://doi.org/10.1016/j.cie.2022.108603}.
\newblock URL
  \url{https://www.sciencedirect.com/science/article/pii/S0360835222005988}.

\bibitem[van Essen et~al.(2014)van Essen, Bosch, Hans, van Houdenhoven, and
  Hurink]{vanEssen2014groupSched}
J.~Theresia van Essen, Jo{\"e}l~M. Bosch, Erwin~W. Hans, Mark van Houdenhoven,
  and Johann~L. Hurink.
\newblock Reducing the number of required beds by rearranging the or-schedule.
\newblock \emph{OR Spectrum}, 36\penalty0 (3):\penalty0 585--605, Jul 2014.
\newblock ISSN 1436-6304.
\newblock \doi{10.1007/s00291-013-0323-x}.
\newblock URL \url{https://doi.org/10.1007/s00291-013-0323-x}.

\bibitem[{Van Riet} and Demeulemeester(2015)]{vanriet2015tradElectiveReview}
Carla {Van Riet} and Erik Demeulemeester.
\newblock Trade-offs in operating room planning for electives and emergencies:
  A review.
\newblock \emph{Operations Research for Health Care}, 7:\penalty0 52--69, 2015.
\newblock ISSN 2211-6923.
\newblock \doi{https://doi.org/10.1016/j.orhc.2015.05.005}.
\newblock URL
  \url{https://www.sciencedirect.com/science/article/pii/S2211692314200646}.
\newblock ORAHS 2014 - The 40th international conference of the EURO working
  group on Operational Research Applied to Health Services.

\bibitem[\v{C}ern\'y(1985)]{Cerny1985}
V.~\v{C}ern\'y.
\newblock Thermodynamical approach to the traveling salesman problem: An
  efficient simulation algorithm.
\newblock \emph{Journal of Optimization Theory and Applications}, 45\penalty0
  (1):\penalty0 41--51, 1985.

\bibitem[Watkins and Dayan(1992)]{watkins1992qlearning}
Christopher J. C.~H. Watkins and Peter Dayan.
\newblock Q-learning.
\newblock \emph{Machine Learning}, 8\penalty0 (3):\penalty0 279--292, May 1992.
\newblock ISSN 1573-0565.
\newblock \doi{10.1007/BF00992698}.
\newblock URL \url{https://doi.org/10.1007/BF00992698}.

\bibitem[Xiang et~al.(2015)Xiang, Yin, and Lim]{xiang2015antColoorsp}
Wei Xiang, Jiao Yin, and Gino Lim.
\newblock An ant colony optimization approach for solving an operating room
  surgery scheduling problem.
\newblock \emph{Computers \& Industrial Engineering}, 85:\penalty0 335--345,
  2015.
\newblock ISSN 0360-8352.
\newblock \doi{https://doi.org/10.1016/j.cie.2015.04.010}.
\newblock URL
  \url{https://www.sciencedirect.com/science/article/pii/S036083521500159X}.

\bibitem[Yazdi et~al.(2020)Yazdi, Zandieh, and Haleh]{yazdi2020elective}
M.~Yazdi, M.~Zandieh, and H.~Haleh.
\newblock A mathematical model for scheduling elective surgeries for minimizing
  the waiting times in emergency surgeries.
\newblock \emph{International Journal of Engineering}, 33\penalty0
  (3):\penalty0 448--458, 2020.
\newblock ISSN 1025-2495.
\newblock \doi{10.5829/ije.2020.33.03c.09}.
\newblock URL \url{https://www.ije.ir/article_104670.html}.

\bibitem[Zhu et~al.(2020)Zhu, Fan, Liu, Yang, and Pardalos]{zhu2020dyn3Stage}
Shuwan Zhu, Wenjuan Fan, Tongzhu Liu, Shanlin Yang, and Panos~M. Pardalos.
\newblock Dynamic three-stage operating room scheduling considering patient
  waiting time and surgical overtime costs.
\newblock \emph{Journal of Combinatorial Optimization}, 39\penalty0
  (1):\penalty0 185--215, Jan 2020.
\newblock ISSN 1573-2886.
\newblock \doi{10.1007/s10878-019-00463-5}.
\newblock URL \url{https://doi.org/10.1007/s10878-019-00463-5}.

\end{thebibliography}

\appendix

\section{Model bounded by Beds} \label{sec:relaxed_model_bed}

Formulation \eqref{mod1:FO}-\eqref{mod1:rest4} assumes that beds are the scheduling bottleneck. Our only decision variable $x_{rk}$ is a binary one that indicates whether surgery $k$ is allocated to bed $r$ ($x_{rk} = 1$) or not ($x_{rk} = 0$).

The objective function \eqref{mod1:FO} minimizes the best case makespan, Constraints \eqref{mod1:rest1} calculate the minimal allocated time for each room, where $T^{T}_{k}:= \sum_{t \in T_k} \gamma_{t}^{d} + \gamma_{t}^{m}$ is the minimal total time the patient from surgery $s$ takes from arriving at the hospital until discharge (sum of duration $\gamma_{t}^{d}$ and moving $\gamma_{t}^{m}$ times), Constraints \eqref{mod1:rest2} ensure each surgery is allocated to some room. Constraint \eqref{mod1:rest3} allocates an arbitrary surgery to a room for symmetry breaking, we display the surgery $0$ being allocated to bed $0$,  and Constraints \eqref{mod1:rest4} state all $x_{rk}$ variables as binary.

\vspace{-0.7cm}
\begin{flalign}
    \indent & Minimize \ z_1  && \label{mod1:FO}
\end{flalign}
\indent Subject to:
\begin{flalign}
    \indent & \qquad \sum_{k \in K} T^{T}_{k}x_{rk} \leq z_1   && \forall \ r \in R^{bed} \label{mod1:rest1} \\ 
    \indent & \qquad \sum_{r \in R^{bed}} x_{rk} = 1            && \forall \ k \in K \label{mod1:rest2} \\ 
    \indent & \qquad  x_{00} = 1   &&            \label{mod1:rest3} \\                                           
    \indent & \qquad x_{rk} \in \{0,1\} &&  \forall \ r \in R^{bed} \quad \forall \ k \in K \label{mod1:rest4}
\end{flalign}

\section{Model bounded by Operation Rooms} \label{sec:relaxed_model_or}

The other proposed solution, Formulation \eqref{mod2:FO}-\eqref{mod2:rest6}, also offers a relaxed approach to solving the IORSP. It assumes that ORs are the scheduling bottleneck. Our objective function, \eqref{mod2:FO}, minimizes the hypothetical makespan. Constraints \eqref{mod2:rest1} estimate the best possible makespan for each OR, where $T^{F}_{k}:= \gamma_{First(k)}^{d}+ \gamma_{First(k)}^{m}$ is the duration and moving times before the OR  for surgery $k$, $T^{S}_{k}$ is the surgery added to the moving time after it and $T^{L}_{k}$ is the sum of all duration and transfer times after the OR task. Analogous to these constants, we have our binary decision variables, $f_{rk}$ indicating if surgery $k$ is the first one allocated to OR $r$, $x_{rk}$ indicates if surgery $k$ is allocated to OR $r$ and $l_{rk}$ indicates if surgery $k$ is the last one allocated to OR $r$.

Constraints \eqref{mod2:rest2} ensure that all surgeries are allocated to a room, while Constraints \eqref{mod2:rest3} and \eqref{mod2:rest4} ensure that each OR has a first and last scheduled surgery. Constraints \eqref{mod2:rest5} ensure that surgery $k$ can only be the first or last if it is allocated to the respective OR, while Constraints \eqref{mod2:rest6} define all $x_{rk}$, $f_{rk}$, and $l_{rk}$ variables as binary.

\begin{flalign}
    \indent & Minimize \ z_2 && \label{mod2:FO}
\end{flalign}
\indent Subject to:
\begin{flalign}
    \indent & \qquad \sum_{k \in K_{r}} (T^{F}_{k}f_{rk} + T^{S}_{k}x_{rk} + T^{L}_{k}l_{rk}) \leq z_2   && \forall \ r \in R^{or}  \label{mod2:rest1} \\
    \indent & \qquad \sum_{r \in R^{or}_{k}} x_{rk} = 1            && \forall \ k \in K \label{mod2:rest2} \\
    \indent & \qquad \sum_{k \in K_{r}} f_{rk} = 1            && \forall \ r \in R^{or} \label{mod2:rest3} \\
    \indent & \qquad \sum_{k \in K_{r}} l_{rk} = 1            && \forall \ r \in R^{or} \label{mod2:rest4} \\
   \indent & \qquad x_{rk} \geq f_{rk} + l_{rk} &&  \forall \ r \in R^{or} \quad \forall \ k \in K_{r}  \label{mod2:rest5} \\
    \indent & \qquad x_{rk},f_{rk},l_{rk} \in \{0,1\} &&  \forall \ r \in R^{or} \quad \forall \ k \in K_{r}  \label{mod2:rest6}
\end{flalign}

\section{Lower bound computations} \label{sec:relaxed_model_results}

\autoref{tab:resultLiteratureForm} shows the results of our proposed models with the literature instances. For each instance is shown its name, number of surgeries, and number of rooms sorted by room type (for instance ``CASE\_10'', using our nomenclature, there are $5$ beds, $4$ ORs, $5$ ICUs, and $20$ wards) and the computed lower bound and CPU time spent for each model. 

Analysing the results, the model bounded by ORs (\ref{mod2:FO}-\ref{mod2:rest6}) is much easier to solve than that bounded by Beds (\ref{mod1:FO}-\ref{mod1:rest4}). In contrast, only seven out of the $24$ instances have optimal lower bounds with 3600 seconds, while all computed lower bounds by the OR model are optimal. Our statistical testing showed the Model Bed to generate superior bounds with $p$-value $=0.0000035$. Thus, the instances are more bounded by Bed availability by a large margin.

\begin{table}[!h]
\centering
\caption{Literature instances formulation results.}
\label{tab:resultLiteratureForm}
\scalebox{0.65}{
\begin{tabular}{lrr|rr|rr} 
Model      &             &            & \multicolumn{2}{c|}{Bed} & \multicolumn{2}{c}{OR} \\ \hline
Instance   & Surgeries   & Rooms      & Lower Bound       & Time(s)    & Lower Bound     & Time(s)    \\ \hline
CASE\_01   & 50          & 5,2,5,10   & \textbf{16.41}    & 3600.01    & 4.57            & 0.01       \\
CASE\_02   & 100         & 5,2,5,10   & \textbf{35.32}    & 340.21     & 7.89            & 0.01       \\
CASE\_03   & 150         & 5,2,5,10   & \textbf{49.06}    & 243.07     & 11.69           & 0.01       \\
CASE\_04   & 200         & 5,2,5,10   & \textbf{67.05}    & 355.09     & 14.93           & 0.01       \\
CASE\_05   & 50          & 5,3,5,10   & \textbf{16.88}    & 3600.01    & 3.72            & 0.02       \\
CASE\_06   & 100         & 5,3,5,10   & \textbf{32.94}    & 145.64     & 5.97            & 0.02       \\
CASE\_07   & 150         & 5,3,5,10   & \textbf{48.81}    & 95.60      & 7.89            & 0.03       \\
CASE\_08   & 200         & 5,3,5,10   & \textbf{67.50}    & 281.53     & 10.09           & 0.02       \\
CASE\_09   & 50          & 5,4,5,20   & \textbf{7.60}     & 3600.01    & 2.85            & 0.10       \\
CASE\_10   & 100         & 5,4,5,20   & \textbf{16.53}    & 3600.02    & 4.64            & 0.02       \\
CASE\_11   & 150         & 5,4,5,20   & \textbf{24.31}    & 3600.01    & 5.92            & 0.06       \\
CASE\_12   & 200         & 5,4,5,20   & \textbf{33.60}    & 3600.02    & 7.73            & 0.04       \\
CASE\_13   & 50          & 5,5,5,20   & \textbf{8.11}     & 3600.01    & 2.67            & 0.19       \\
CASE\_14   & 100         & 5,5,5,20   & \textbf{16.92}    & 3600.01    & 3.91            & 0.07       \\
CASE\_15   & 150         & 5,5,5,20   & \textbf{25.50}    & 3600.01    & 5.10            & 0.10       \\
CASE\_16   & 200         & 5,5,5,20   & \textbf{35.27}    & 3600.01    & 6.66            & 0.04       \\
CASE\_17   & 50          & 5,10,5,20  & \textbf{9.03}     & 3600.01    & 2.44            & 65.02      \\
CASE\_18   & 100         & 5,10,5,20  & \textbf{16.99}    & 3600.01    & 2.70            & 50.53      \\
CASE\_19   & 150         & 5,10,5,20  & \textbf{26.23}    & 3600.01    & 3.29            & 0.64       \\
CASE\_20   & 200         & 5,10,5,20  & \textbf{33.55}    & 3600.01    & 3.96            & 19.58      \\
CASE\_21   & 50          & 5,10,5,30  & \textbf{6.09}     & 0.51       & 2.36            & 76.05      \\
CASE\_22   & 100         & 5,10,5,30  & \textbf{11.05}    & 3600.16    & 2.62            & 15.84      \\
CASE\_23   & 150         & 5,10,5,30  & \textbf{15.70}    & 3600.02    & 3.27            & 160.01     \\
CASE\_24   & 200         & 5,10,5,30  & \textbf{21.26}    & 3600.01    & 3.90            & 95.01      \\ \hline
\multicolumn{3}{l|}{Average times}         &               & 3450.01    &           & 43.04      \\
\multicolumn{3}{l|}{Averages RPD(\%)} &    0.000                &     &     77.321            &    
\end{tabular}}
\end{table}

\begin{figure}[h]
    \centering
    \scalebox{0.32}{\includegraphics{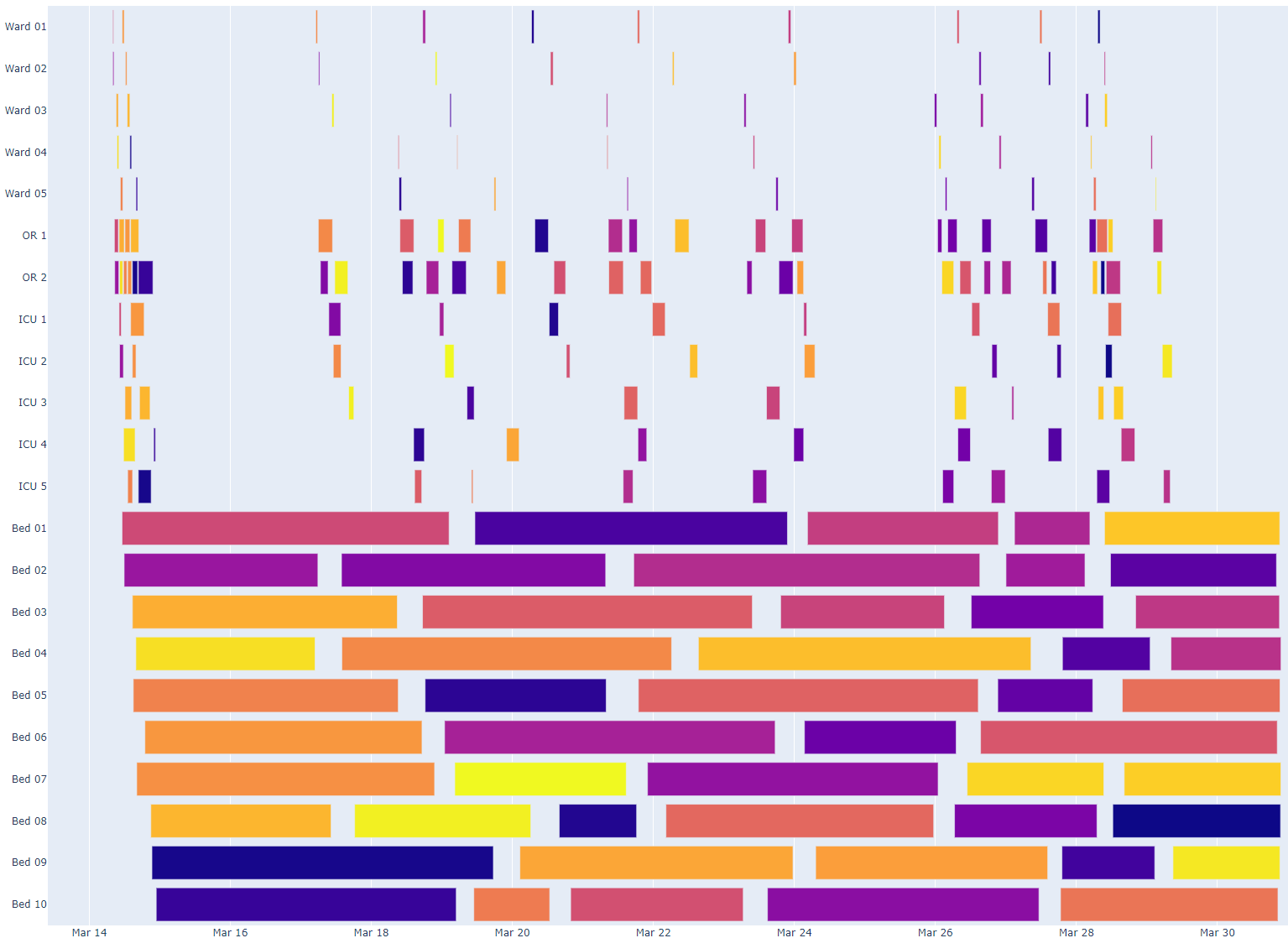}}
    \caption{Example of a solution found by BRKGA-QL for a literature instance bounded by bed availability.}
    \label{fig:solutionLit}
\end{figure}

 We can visualise this attribute in \autoref{fig:solutionLit}, a heuristic solution of ``CASE\_01''. Each unique colour represents a patient, and we can observe that all rooms, except for beds, have sparse allocations. A bed is reserved from the moment the patient enters a Ward. This explains the empty spaces between the bed usages; however, they are fully utilized.

\autoref{tab:resultCaseForm} shows the results of our proposed models with the case study instances. For each instance, the instances' names, number of surgeries, number of rooms sorted by room type, the computed lower bound in days, and CPU time spent for each model type are presented. Analyzing the results, we can observe that $16$ instances were more bounded by OR availability and four by Bed availability. The statistical testing shows that the OR formulation computed better bounds with $p$-value $=0.005162$. This shows some balance during the instances' procedural generation. As in the literature instances, the model bounded by ORs is much easier to solve than the model bounded by Beds, whereas only nine out of the $20$ instances have optimal lower bounds within $3600$ seconds, and all computed lower bounds by the OR model are optimal.

\begin{table}[!h]
\centering
\caption{Case study instances formulation results.}
\label{tab:resultCaseForm}
\scalebox{0.70}{
\begin{tabular}{lrr|rr|rr}
Model      &             &            & \multicolumn{2}{c|}{Bed} & \multicolumn{2}{c}{OR} \\ \hline
Instance   & Surgeries   & Rooms      & Lower Bound       & Time(s)    & Lower Bound      & Time(s)   \\ \hline
p070       & 58          & 31,3,2,4   & 3.899             & 0.32       & \textbf{4.440}   & 0.01      \\
p078       & 66          & 23,3,3,4   & \textbf{5.536}    & 3600.02    & 4.722            & 0.07      \\
p087       & 75          & 29,3,3,4   & \textbf{5.247}    & 3600.02    & 5.146            & 0.06      \\
p093       & 81          & 30,3,3,4   & 5.253             & 3600.01    & \textbf{5.485}   & 0.07      \\
p098       & 86          & 40,3,2,4   & 4.444             & 26.32      & \textbf{5.719}   & 0.02      \\
p100       & 84          & 39,4,3,5   & 4.486             & 3600.03    & \textbf{4.604}   & 0.14      \\
p109       & 93          & 35,4,3,5   & \textbf{5.364}    & 3600.01    & 5.015            & 0.36      \\
p120       & 104         & 46,4,4,5   & 4.699             & 3600.03    & \textbf{5.466}   & 0.46      \\
p138       & 118         & 43,5,4,5   & \textbf{5.224}    & 3600.01    & 4.945            & 0.04      \\
p153       & 133         & 71,5,4,7   & 3.823             & 2.32       & \textbf{5.486}   & 0.09      \\
p165       & 141         & 78,6,4,7   & 3.895             & 4.76       & \textbf{5.019}   & 5.69      \\
p178       & 154         & 88,6,4,6   & 3.848             & 2.89       & \textbf{5.335}   & 1.09      \\
p183       & 159         & 89,6,4,6   & 3.875             & 282.51     & \textbf{5.556}   & 1.76      \\
p189       & 161         & 71,7,5,7   & 4.849             & 2912.48    & \textbf{4.902}   & 8.45      \\
p192       & 164         & 72,7,4,8   & 4.701             & 3600.01    & \textbf{4.964}   & 3.97      \\
p193       & 165         & 78,7,5,7   & 4.113             & 3600.02    & \textbf{4.952}   & 13.92     \\
p197       & 173         & 99,6,5,7   & 3.829             & 30.11      & \textbf{5.767}   & 3.16      \\
p201       & 177         & 73,6,4,8   & 4.875             & 3600.03    & \textbf{6.053}   & 5.82      \\
p216       & 188         & 95,7,5,9   & 3.842             & 24.05      & \textbf{5.505}   & 3.19      \\
p233       & 197         & 88,9,4,10  & 4.484             & 3600.08    & \textbf{4.778}   & 36.02     \\ \hline
\multicolumn{3}{l|}{Averages}         &                  & 2144.302   &            & 4.220     \\
\multicolumn{3}{l|}{Averages RDP(\%)} & 13.944          &            &     1.424   &        
\end{tabular}}
\end{table}

\begin{figure}[!h]
    \centering
    \scalebox{0.32}{\includegraphics{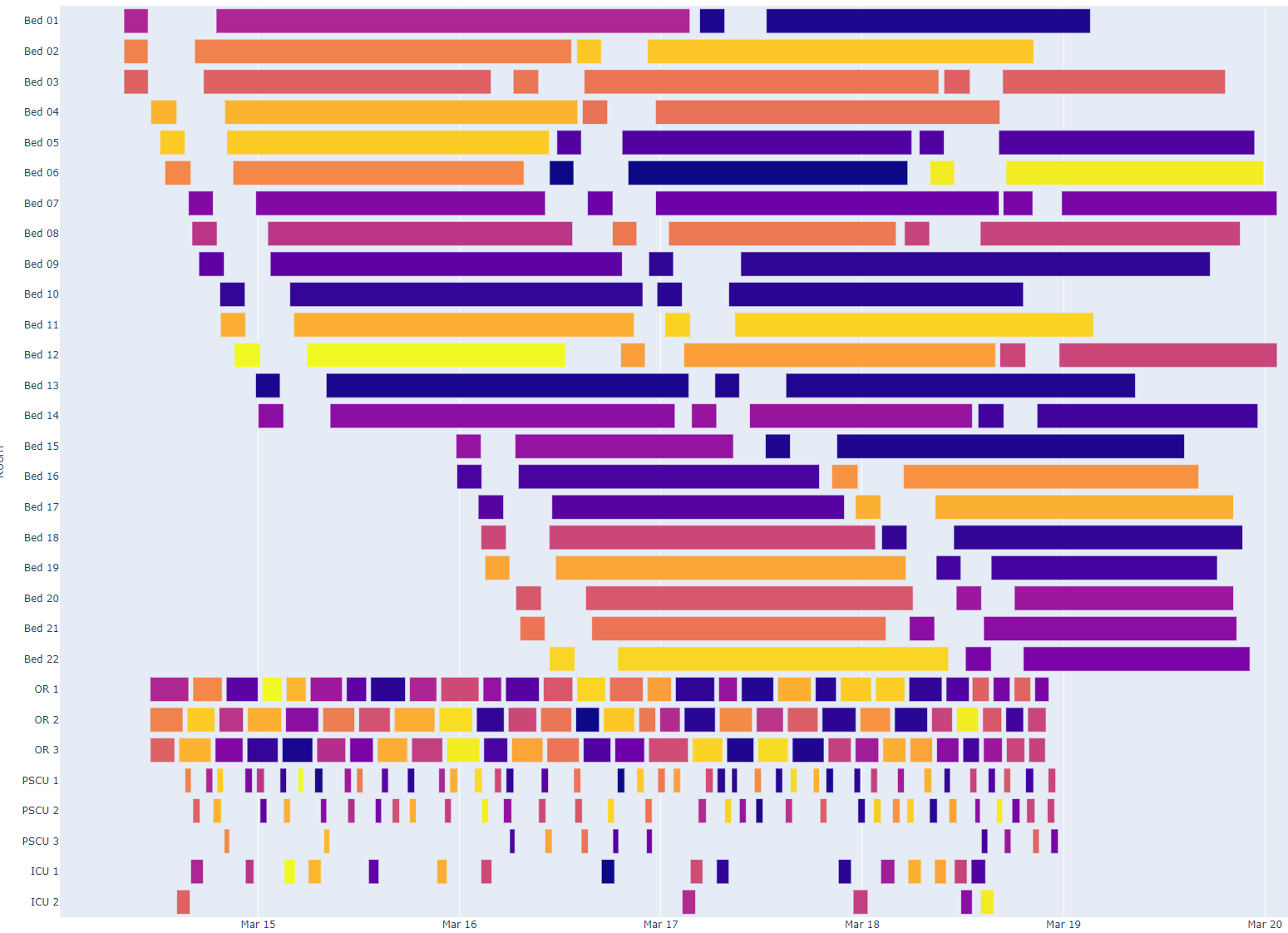}}
    \caption{Example of a solution found by BRKGA-QL for a case study instance bounded by OR availability.}
    \label{fig:solutionCaseS}
\end{figure}

To visualize how a solution with all the availability slots looks like, \autoref{fig:solutionCaseS} shows the Gantt view of the best solution of instance $p098$ without availability slots. We can observe that the operating rooms are allocated very close to the limit of their capacity, and the last occupied beds are released almost simultaneously, not as well distributed as in \autoref{fig:solutionLit}.

\end{document}